\documentclass{article} 
\usepackage{jvt,times}

\usepackage{amsmath,amsfonts,bm}

\def\eqref#1{equation~\ref{#1}}

\def\1{\bm{1}}

\def\ri{{\textnormal{i}}}

\DeclareMathAlphabet{\mathsfit}{\encodingdefault}{\sfdefault}{m}{sl}
\SetMathAlphabet{\mathsfit}{bold}{\encodingdefault}{\sfdefault}{bx}{n}

\usepackage{hyperref}
\usepackage{url}
\usepackage{graphicx}
\usepackage{tikz}
\usepackage{wrapfig}
\usepackage{needspace}
\usepackage{booktabs}
\usepackage{tabularx}
\usepackage{multirow}
\usepackage{xspace}
\usepackage[dvipsnames]{xcolor}
\usepackage{pifont}

\title{JVT: Just Video Transformer makes \\ pixel-space video diffusion simple}

\author{
  Jianjun Zhou$^{1,2,3}$\thanks{Equal Contribution.} \quad
  Hua Geng$^{1}$\footnotemark[1] \quad
  Haoyu Guo$^{3}$\thanks{Corresponding Author. Email: \texttt{guohaoyu@pjlab.org.cn}} \\
  $^1$Zhejiang University \quad
  $^2$Shanghai Innovation Institute \quad
  $^3$Shanghai AI Laboratory \\
  \smallskip
  {\small \color{RoyalBlue}{\texttt{https://zhoutimemachine.github.io/JVT}}}
}

\newcommand{\cmark}{\ding{51}}
\newcommand{\xmark}{\ding{55}}

\hypersetup{
    colorlinks=true,
    linkcolor=black,
    citecolor=black,
    urlcolor=RoyalBlue
}

\begin{document}

\maketitle
\begin{figure}[htbp]
    \centering
    \vspace{-2em}
    \includegraphics[width=\linewidth]{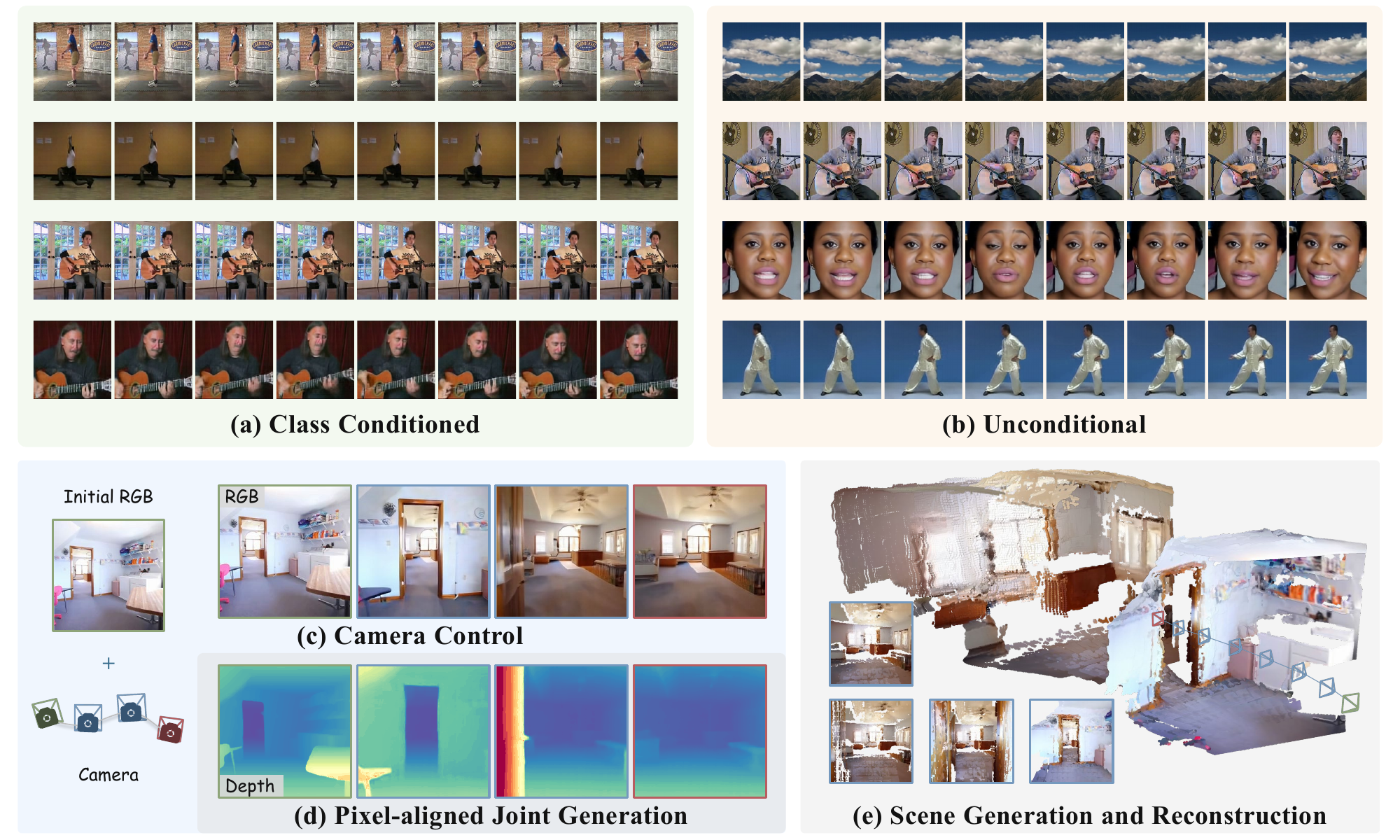}
    \caption{JVT generates videos directly in pixel space with a simple DiT-style transformer, supporting (a) class-conditioned and (b) unconditional generation. Given an initial RGB frame and a target camera trajectory, it jointly generates (c) RGB frames and disparity, visualized as (d) depth maps. These outputs yield (e) colored point clouds through geometric unprojection, without an additional reconstruction model.}
    \label{fig:teaser}
\end{figure}

\begin{abstract}
    Direct diffusion modeling of high-dimensional raw video data is challenging, and existing pixel-space methods often rely on specialized architectures. We introduce \textbf{JVT} (\textbf{J}ust \textbf{V}ideo \textbf{T}ransformer), a simple large-patch DiT-style Transformer for pixel-space video diffusion, without a pretrained autoencoder, auxiliary detailers, cascaded stages, or U-Net-style skip connections. We systematically examine video patch embeddings, register tokens, and adaptive layer normalization to identify design choices that balance generation quality and computational efficiency. The resulting JVT design achieves generation quality competitive with latent diffusion baselines in both class-conditioned and unconditional settings. The pixel-space formulation of JVT also naturally supports joint generation of pixel-aligned modalities. We demonstrate this capability through camera-conditioned RGB-disparity generation, which improves camera following relative to RGB-only baselines and yields explicit scene geometry without an additional reconstruction model at inference. These results highlight the potential of simple DiT-style Transformers for pixel-space video diffusion, and we hope our findings offer useful insights and practical guidance for future research.
\end{abstract}

\section{Introduction}

Diffusion and flow-based models have become leading paradigms for visual generative modeling, driven by rapid advances in algorithm designs and model architectures. Early diffusion models~\citep{ho2020ddpm,dhariwal2021adt} operate directly on pixels with classical U-Nets~\citep{ronneberger2015unet}, but struggle with efficiency and fidelity at high resolutions. To alleviate this difficulty, latent diffusion moves denoising from pixel space to a compact latent space produced by a separately trained autoencoder~\citep{rombach2022ldm}. Combined with the Diffusion Transformer~(DiT) architecture~\citep{peebles2023dit}, this formulation has become a standard design for high-resolution visual generation.

Despite its efficiency, latent diffusion is inherently bounded by the autoencoder, including reconstruction error~\citep{yu2026pixeldit,zheng2026rae}, additional training and inference costs~\citep{leng2025repae,cheng2025leanvae,wang2026eliminating}, and limited cross-modal capability for joint modeling~\citep{bounoua2024mldm,qiu2025mvae}. Thus recent works revisit pixel-space diffusion and demonstrate competitive quality on image generation~\citep{li2026jit,yu2026pixeldit,lu2026pmf,wang2026minit2i}. In contrast, video generation is still dominated by latent DiTs~\citep{ma2025latte,kong2024hunyuanvideo,yang2025cogvideox,wan2025wan21}, while pixel-space models mostly retain U-Net-style skip connections~\citep{gu2024matryoshka,bartal2024lumiere,hoogeboom2025sid2} or cascaded generation~\citep{menapace2024snapvideo,skorokhodov2024hpd}. This motivates us to investigate how far a simple DiT-style Transformer can go in direct pixel-space video generation.

We introduce JVT, a pixel-space video generative model built on a simple DiT-style Transformer backbone, without detailers, cascaded stages or U-Net-style skip connections. Following JiT~\citep{li2026jit}, JVT predicts clean videos from noisy pixels under a rectified-flow objective. To make raw videos tractable, we explore variants of video patch embedding for efficient spatiotemporal downsampling. We further examine the design choices of register tokens~\citep{darcet2024registers} and adaptive layer normalization (AdaLN)~\citep{peebles2023dit}, which can carry noise level or class condition and amplify it by register guidance~\citep{starodubcev2026registerguidance} or classifier-free guidance~\citep{ho2021cfg}. With these design choices, JVT achieves generation quality competitive with latent diffusion baselines in both class-conditioned and unconditional settings.

Pixel-space video modeling also provides a direct interface for jointly generating video and pixel-aligned modalities, which typically requires reusing an RGB-specific autoencoder in latent diffusion~\citep{ke2024marigold,zhu2025aether,huang2025unityvideo}. We extend JVT to camera-conditioned joint RGB-disparity generation to demonstrate this cross-modal capability. Without an additional reconstruction model at inference, the generated disparity yields a sequence of point clouds in world coordinates, which enables explicit 4D world modeling. Experiments further show that joint RGB-disparity generation improves camera following relative to RGB-only JVT.

The main contributions of this work are as follows:
\begin{itemize}
    \item We present JVT, the first pixel-space video diffusion model built on a simple DiT-style Transformer backbone, without detailers, cascaded stages, or U-Net-style skip connections.
    \item We systematically study video patch embeddings and conditioning mechanisms, identifying designs that improve the quality-efficiency trade-off. In class-conditioned and unconditional settings, JVT achieves generation quality competitive with latent diffusion baselines.
    \item We formulate pixel-aligned joint generation as a native extension of JVT. We further demonstrate this capability through camera-conditioned RGB-disparity video generation, which enables explicit 4D scene representation and improves camera following relative to RGB-only generation.
\end{itemize}

\section{Related Work}

\noindent\textbf{High-resolution pixel-space diffusion.}
Pixel-space diffusion modeling for high-resolution visual data has long relied on specialized architectures, including cascaded diffusion~\citep{ho2022cascaded,teng2024relay}, cross-resolution designs~\citep{gu2024matryoshka,chen2025pixelflow}, multiscale or U-shaped backbones~\citep{hoogeboom2023sid,crowson2024hdit,hoogeboom2025sid2,he2026hyperdit,guo2026pixelu}, frequency decomposition~\citep{ma2026deco,lin2026frepix,ma2026frequencybooster}, and auxiliary detail modules~\citep{wang2026pixnerd,chen2026dip,yu2026pixeldit}. Though effective, these approaches often introduce additional architectural or training complexity, which can limit efficiency and scalability. Recently, JiT~\citep{li2026jit} achieves high-resolution image generation competitive with latent diffusion using a simple large-patch DiT-style Transformer. This paradigm is also extended to one-step image generation~\citep{lu2026pmf} and text-to-image generation~\citep{wang2026minit2i,jiang2026empirical}, further demonstrating the effectiveness.

\noindent\textbf{Video Transformer architectures.}
Modern video Transformers largely retain the standard blocks of 2D Vision Transformers~(ViTs), making patch embedding the key adaptation to video inputs~\citep{tong2022videomae,wang2023videomaev2,bardes2024vjepa,assran2025vjepa2}. A common choice is tubelet embedding~\citep{arnab2021vivit}, which encodes non-overlapping spatiotemporal patches into tokens via 3D convolution. This architecture is also widely adopted in latent video DiTs, whereas pixel-space video diffusion methods still typically rely on U-shaped architectures~\citep{gu2024matryoshka,bartal2024lumiere,hoogeboom2025sid2} or cascaded generation~\citep{menapace2024snapvideo,skorokhodov2024hpd}. Our work aims to explore whether a simple DiT-style video Transformer can also support direct pixel-space video generation.

\section{Method}

\subsection{Preliminaries}

Though denoising diffusion models commonly use noise, score or velocity prediction~\citep{dhariwal2021adt,song2021sgm,karras2022edm,salimans2022vpred,lipman2023flowmatching,liu2023rectifiedflow}, JiT proposes clean data prediction and velocity loss supervision for pixel-space diffusion~\citep{li2026jit}. Following this formulation, JVT predicts clean video $\hat{x}_0$ from noisy pixels $x_t$:
\begin{equation}
    x_t=(1-t)x_0+t\epsilon,\qquad
    \hat{x}_0=f_\theta(x_t,t,c),\qquad
    \hat{v}=(\hat{x}_0-x_t)/t,
    \label{eq:clean-video-prediction}
\end{equation}
where noise $\epsilon\sim\mathcal{N}(0,I)$, noise level $t\in(0,1]$, and $c$ denotes the available conditioning. Thus the standard rectified-flow velocity loss can be represented as:
\begin{equation}
    \mathcal{L}_{\mathrm{MSE}}
    =\mathbb{E}\left\|\hat{v}-v\right\|_2^2
    =\mathbb{E}\|(\hat{x}_0-x_0)/t\|_2^2,\qquad
    v=x_0-\epsilon=(x_0-x_t)/t.
    \label{eq:rectified-flow-loss}
\end{equation}

\begin{figure}[h]
    \centering
    \includegraphics[width=\linewidth]{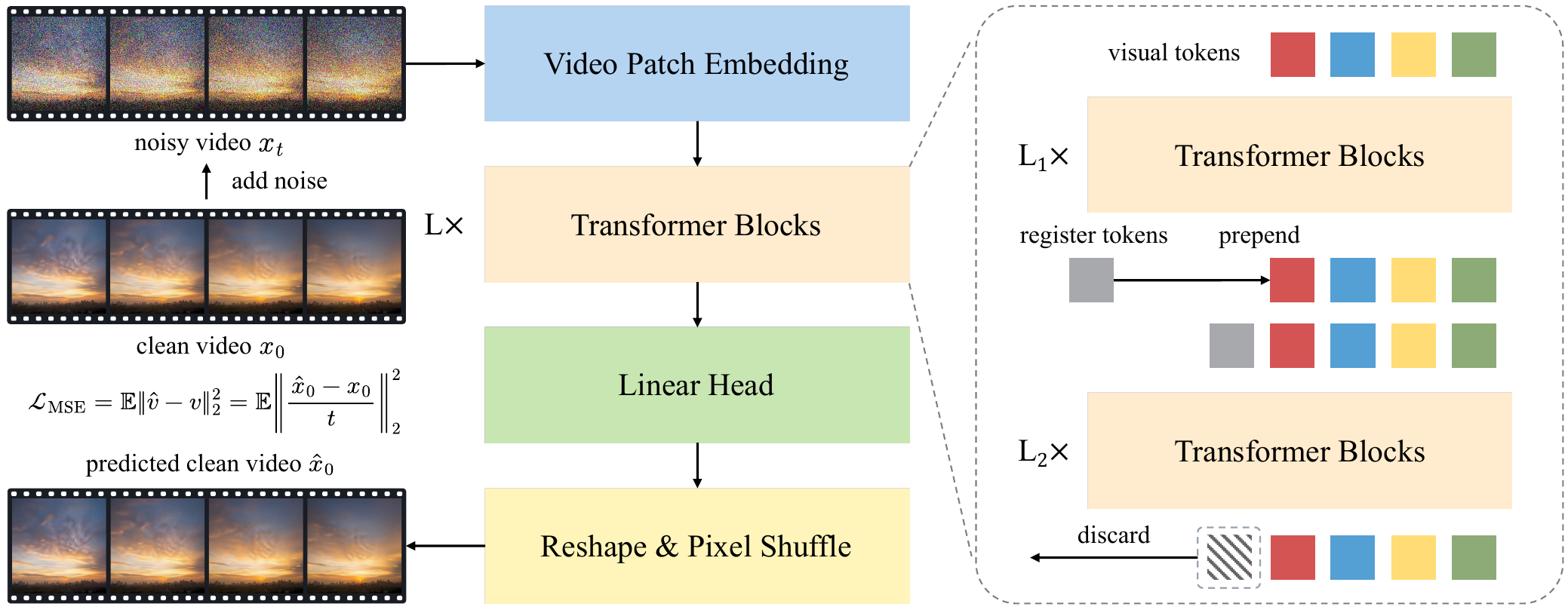}
    \caption{\textbf{JVT architecture and register token design}. The network predicts a clean video $\hat{x}_0$ from a noisy input $x_t$. Register tokens are prepended to visual tokens after the first $L_1$ blocks, and then discarded after the remaining $L_2$ blocks, where $L=L_1+L_2$.}
    \label{fig:transformer-and-register}
\end{figure}

\subsection{Patch Embedding}\label{sec:method_patch_embedding}

Patch embedding is central to pixel-space DiTs, as it provides the initial downsampling that latent DiTs obtain from an autoencoder. For videos, it also aggregates spatiotemporal information while controlling the token sequence length.

\begin{figure}[h]
    \centering
    \includegraphics[width=\linewidth]{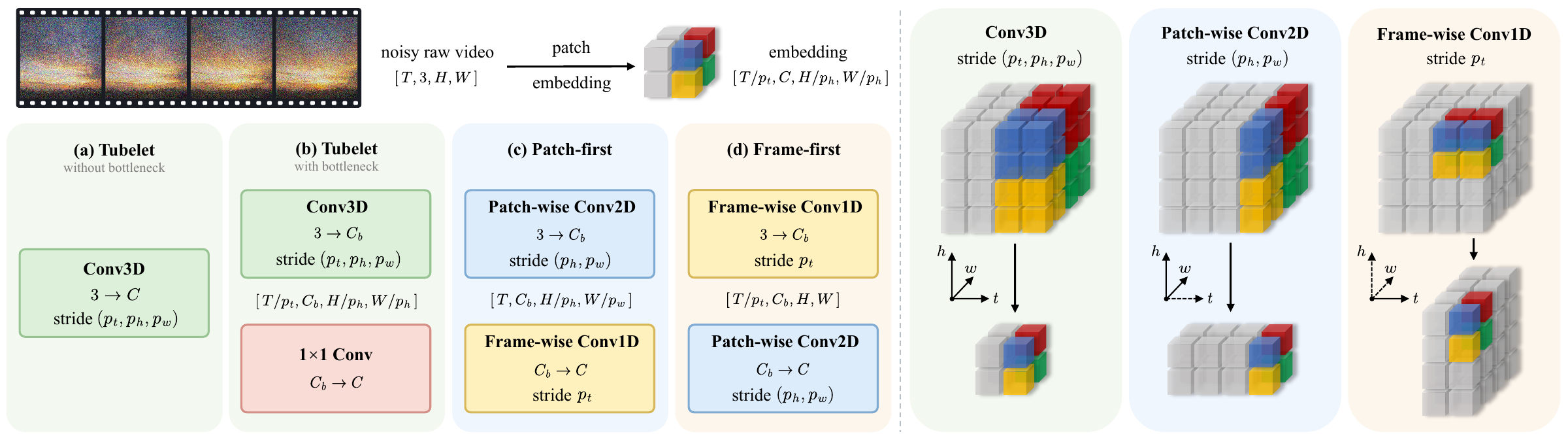}
    \caption{\textbf{Video patch embedding variants (left) and their convolutional components (right)}. Assume non-overlapping patches with size $p_t\times p_h\times p_w$. $T$, $H$, $W$ are divisible by $p_t$, $p_h$, $p_w$, respectively.}
    \label{fig:3dpatchify}
\end{figure}

Following standard video Transformer designs, we use plain tubelet embedding as our baseline. As illustrated in Fig.~\ref{fig:3dpatchify}~(a), it uses a single 3D convolution to project each non-overlapping patch of temporal size $p_t$ and spatial size $p_h\times p_w$ into a $C$-dimensional token.

\noindent\textbf{Bottleneck structure.}
Following JiT~\citep{li2026jit}, we introduce a channel bottleneck into tubelet embedding (Fig.~\ref{fig:3dpatchify}~(b)). A 3D convolution first projects each patch into a $C_b$-dimensional representation, which is then expanded to $C$ channels by a pointwise projection. We typically choose $C_b<C$ to encourage compact, low-dimensional representations~\citep{tishby2000ib,rifai2011contractiveae,makhzani2013kae,alemi2017deepvib}.

\noindent\textbf{Spatiotemporal factorization.}
Inspired by the factorization principle of depthwise separable convolution~\citep{chollet2017xception,howard2017mobilenets0}, we separate the spatial and temporal operations of tubelet embedding. Specifically, a patch-wise 2D convolution extracts spatial features within each frame, while a frame-wise 1D convolution aggregates temporal information at each spatial location. Applying the spatial operation first yields patch-first embedding (Fig.~\ref{fig:3dpatchify}~(c)), while reversing the order yields frame-first embedding (Fig.~\ref{fig:3dpatchify}~(d)).

\subsection{Register Tokens}\label{sec:method_register}
Register tokens provide additional computational slots in ViTs~\citep{darcet2024registers} and also support in-context conditioning and improve generation quality in pixel-space image DiTs~\citep{li2026jit,starodubcev2026registerguidance}. We extend this mechanism to JVT and examine its roles in conditioning and sampling guidance.

As illustrated in Fig.~\ref{fig:transformer-and-register}, given $L=L_1+L_2$ Transformer blocks, register tokens are prepended to the visual sequence after the first $L_1$ blocks and discarded after the remaining $L_2$ blocks. Let $X^{(L_1)}\in\mathbb{R}^{N\times C}$ denote the $N$ visual tokens at the insertion point and $R_0\in\mathbb{R}^{\ell_r\times C}$ the $\ell_r$ learnable base registers. With $e_t,e_y\in\mathbb{R}^{C}$ denoting the noise level and class embeddings, respectively, we construct the register tokens and augmented sequence as
\begin{equation}
    R=R_0+\operatorname{Repeat}(e_r,\ell_r),\qquad
    \widetilde{X}^{(L_1)}=\operatorname{Concat}(R,X^{(L_1)}),
\end{equation}
where $e_r\in\{0,e_t,e_y,e_t+e_y\}$ and $\operatorname{Repeat}(e_r, \ell_r)$ replicates $e_r$ across the $\ell_r$ register positions. Setting $e_r=0$ preserves $R=R_0$ without adding a conditioning embedding, which is different from removing the registers. For $e_r\in\{e_t, e_y, e_t+e_y\}$, the register tokens provide an additional conditioning pathway alongside AdaLN.

\noindent\textbf{Register guidance.}
We also apply register guidance~(RG)~\citep{starodubcev2026registerguidance} during sampling. For the same noisy input and condition, let $v_r$ and $v_{nr}$ denote the velocity predictions of the same model with and without registers, respectively. RG amplifies the register contribution through
\begin{equation}
    v_{rg}
    = v_r + w_{rg} (v_r - v_{nr}),
    \qquad w_{rg}\geq0.
    \label{eq:register-guidance}
\end{equation}
For conditional generation, let $v_u$ denote the unconditional velocity prediction with registers retained. Combining RG with classifier-free guidance~(CFG)~\citep{ho2021cfg} gives
\begin{equation}
    v_{cfg+rg}
    = v_u
    + w_{cfg}(v_r-v_u)
    + w_{rg}(v_r-v_{nr}).
    \label{eq:register-cfg}
\end{equation}
The scales $w_{cfg}$ and $w_{rg}$ separately weight the CFG and RG terms: setting $w_{rg}=0$ recovers CFG, while $w_{cfg}=1$ recovers RG in Eq.~(\ref{eq:register-guidance}). Unlike the original RG procedure~\citep{starodubcev2026registerguidance}, which randomly drops registers during training, we keep them active throughout training and remove them only to compute $v_{nr}$ during sampling. Our experiments show that RG remains effective without register dropout during training.

\subsection{Pixel-aligned Joint Generation}\label{sec:method_joint_generation}
Pixel-aligned signals about geometry, motion, and semantic structure provide complementary cues for video generation. By channel-wise concatenating RGB frames with these signals~\citep{ke2024marigold}, diffusion models can jointly generate RGB content and spatially aligned auxiliary modalities~\citep{zhu2025aether,chen2025deepverse,huang2025unityvideo}.

Pixel-space modeling allows these modalities to be represented on a shared spatiotemporal grid without an RGB-specific autoencoder. After modality-specific normalization, we concatenate an RGB video $x_{\mathrm{rgb}}\in\mathbb{R}^{T\times3\times H\times W}$ with $M$ pixel-aligned modalities along the channel dimension:
\begin{equation}
    x_0
    = [x_{\mathrm{rgb}};x^{(1)};\ldots;x^{(M)}]
    \in\mathbb{R}^{T\times C_{\mathrm{joint}}\times H\times W},
    \label{eq:pixel-aligned-joint-prediction}
\end{equation}
where $x^{(m)}\in\mathbb{R}^{T\times C_m\times H\times W}$ denotes the $m$-th aligned modality and $C_{\mathrm{joint}}=3+\sum_{m=1}^{M}C_m$ is the total number of channels. These modalities may include depth, surface normals, ray maps, segmentation masks, and optical flow. All modalities share the same noise level and are predicted jointly during denoising.

We instantiate this formulation with RGB and disparity, a reparameterization of depth, for camera-conditioned video generation (Sec.~\ref{sec:camera-conditioned-generation}). At inference, we convert the predicted disparity back to depth and unproject it using the camera parameters. Combined with the generated RGB frames, this yields a sequence of colored point clouds without an additional reconstruction model.
\section{Experiments}

\subsection{Main Results}
\noindent\textbf{Datasets.}
For class-conditioned generation and unconditional generation, we conduct experiments on four datasets: UCF-101~\citep{soomro2012ucf101}, FaceForensics~\citep{rossler2018faceforensics}, SkyTimelapse~\citep{xiong2018skytimelapse}, and Taichi-HD~\citep{siarohin2019taichi}. Following settings of previous works~\citep{skorokhodov2022styleganv,ma2025latte}, we use both training and test splits of UCF-101, and only the training split of the other datasets. We randomly extract 16-frame video clips with a frame interval of 3, and resize all video frames to $256\times 256$ resolution for training.

\noindent\textbf{Evaluation metrics.}
We evaluate generation quality with Fréchet Video Distance (FVD)~\citep{unterthiner2019fvd}, using 2,048 generated samples. For efficiency metrics, we report parameter count, floating point operations (FLOPs), peak allocated GPU memory and inference latency. Additional details about these metrics are given in Appendix~(\ref{sec:appendix_efficiency_results}, \ref{sec:appendix_fvd}).

\noindent\textbf{Baselines.}
Our main baseline is Latte~\citep{ma2025latte}, a representative DiT-based latent diffusion model. Other competing methods include GAN-based methods~\citep{skorokhodov2022styleganv,yu2022digan}, autoregressive models~\citep{yan2021videogpt,wang2024omnitokenizer}, U-Net-based latent diffusion~\citep{he2022lvdm,yu2023pvdm} and DiT-based latent diffusion~\citep{liu2024fvdm}.

\begin{table}[t]
    \centering
    \newcommand{\sgvscore}[1]{\textcolor{gray}{#1\textsuperscript{$\dagger$}}}
    \newcommand{\diganscore}[1]{\textcolor{gray}{#1\textsuperscript{$\ddagger$}}}
    \newcommand{\splitscore}[1]{\textcolor{gray}{#1\textsuperscript{*}}}
    \caption{\textbf{FVD@2048 across video generation benchmarks.} Cond. and Uncond. denote class-conditioned and unconditional generation, respectively. Res. denotes spatial resolution, and all methods use 16-frame clips. Gray scores marked with $\dagger$/$\ddagger$ are taken from StyleGAN-V/DIGAN, and other values are sourced from the original papers. Taichi-HD scores marked with $*$ are trained on both training and test splits, excluded from ranking. JVT with and without ``+RG'' denote JVT with register guidance and JVT without registers, respectively. -- denotes unreported results.}
    \label{tab:fvd-main}
    \small
    \setlength{\tabcolsep}{1.25pt}
    \begin{tabular*}{\linewidth}{@{\extracolsep{\fill}}lcc@{\hspace{5pt}}c@{\hspace{4.5pt}}c@{\hspace{4.5pt}}c@{\hspace{4.5pt}}c@{}}
        \toprule
        \multirow{2}{*}{\textbf{Method}} &
        \multirow{2}{*}{\textbf{Res.}} &
        \textbf{Cond.} &
        \multicolumn{4}{c}{\textbf{Uncond.}} \\
        \cmidrule(lr){3-3} \cmidrule(lr){4-7}
        & & UCF-101 & UCF-101
        & FaceForensics
        & SkyTimelapse
        & Taichi-HD \\
        \midrule
        MoCoGAN~\citep{tulyakov2018mocogan} & $256^2$ & -- & \sgvscore{2886.9} & \sgvscore{124.7} & \sgvscore{206.6} & -- \\
        MoCoGAN-HD~\citep{tian2021mocoganhd} & $256^2$ & -- & \sgvscore{1729.6} & \sgvscore{111.8} & \sgvscore{164.1} & \diganscore{144.7} \\
        VideoGPT~\citep{yan2021videogpt} & $256^2$ & -- & \sgvscore{2880.6} & \sgvscore{185.9} & \sgvscore{222.7} & -- \\
        StyleGAN-V~\citep{skorokhodov2022styleganv} & $256^2$ & -- & 1431.0 & 47.41 & 79.52 & -- \\
        MoStGAN-V~\citep{shen2023mostganv} & $256^2$ & -- & 1380.3 & 39.7 & 65.3 & -- \\
        CogVideo~\citep{hong2023cogvideo} & $160^2$ & 626 & -- & -- & -- & -- \\
        DIGAN~\citep{yu2022digan} & $128^2$ & 465 & 577 & -- & 114.6 & \splitscore{128.1} \\
        DIGAN~\citep{yu2022digan} & $256^2$ & -- & \sgvscore{1630.2} & \sgvscore{62.5} & \sgvscore{83.11} & -- \\
        LVDM~\citep{he2022lvdm} & $256^2$ & -- & 372 & -- & 95.2 & \splitscore{99.0} \\
        PVDM~\citep{yu2023pvdm} & $256^2$ & -- & 343.6 & -- & 55.41 & -- \\
        VIDM~\citep{mei2023vidm} & $128^2$ & -- & -- & -- & -- & \underline{121.9} \\
        VIDM~\citep{mei2023vidm} & $256^2$ & -- & 294.7 & -- & 57.4 & -- \\
        GLOBER~\citep{sun2023glober} & $128^2$ & 151.5 & 239.5 & -- & -- & 124.2 \\
        GLOBER~\citep{sun2023glober} & $256^2$ & 168.9 & 252.7 & -- & 78.1 & -- \\
        OmniTokenizer~\citep{wang2024omnitokenizer} & $256^2$ & 191 & -- & -- & -- & -- \\
        VDT~\citep{lu2024vdt} & $64^2$ & -- & \underline{225.7} & -- & -- & -- \\
        FVDM~\citep{liu2024fvdm} & $256^2$ & 468.23 & -- & 55.01 & 106.09 & 194.61 \\
        Latte~\citep{ma2025latte} & $256^2$ & 477.97 & -- & \underline{34.00} & 59.82 & 159.60 \\
        \midrule
        JVT (Ours) & $256^2$ & \underline{136.5} & 413.5 & \underline{32.86} & \textbf{46.58} & \textbf{119.1} \\
        JVT+RG (Ours) & $256^2$ & \textbf{129.2} & \textbf{215.9} & \textbf{31.80} & \underline{47.80} & 124.3 \\
        \bottomrule
    \end{tabular*}
\end{table}

Tab.~\ref{tab:fvd-main} reports class-conditioned results on UCF-101 and unconditional results on the four datasets. JVT outperforms all competing methods in the table, demonstrating the effectiveness of direct pixel-space video modeling. Additional qualitative results are provided in Fig.~\ref{fig:appendix-uncond-comparison} of the Appendix.

\subsection{Ablation Study}
\label{sec:ablation_study}
\noindent\textbf{Video patch embedding.}
We compare the tubelet, patch-first, and frame-first embeddings introduced in Sec.~\ref{sec:method_patch_embedding} while varying the bottleneck width $C_b$. Under identical training settings, all bottlenecked variants in Tab.~\ref{tab:patch-embedding-bottleneck} achieve lower final training loss than the plain tubelet baseline. With an appropriate choice of $C_b$, tubelet and patch-first embeddings also improve FVD while reducing computational cost. Taking both generation quality and efficiency into consideration, we adopt patch-first embedding with $C_b=32$ as the default configuration for JVT. Further patch embedding ablation details are provided in Appendix~\ref{sec:appendix_ablations}.

\begin{table}[t]
    \centering
    \caption{\textbf{Performance of different video patch embedding variants.} Efficiency metrics are measured on one NVIDIA 4090 GPU. Denoiser rows measure the complete model, while Emb. rows measure the isolated video patch embedding module on a $1\times16\times3\times256\times256$ input. Embedding peak allocated memory
    includes the input and embedding parameters.}
    \label{tab:patch-embedding-bottleneck}
    \small
    \setlength{\tabcolsep}{4pt}
    \renewcommand{\arraystretch}{1.15}
    \resizebox{\linewidth}{!}{
        \begin{tabular}{@{}lcccccccccc@{}}
            \toprule
            \textbf{Embedding}
            & \multicolumn{4}{c}{\textbf{Tubelet}}
            & \multicolumn{3}{c}{\textbf{Patch-first}}
            & \multicolumn{3}{c}{\textbf{Frame-first}} \\
            \cmidrule(lr){2-5} \cmidrule(lr){6-8} \cmidrule(lr){9-11}
            \textbf{$C_b$}
            & - & 32 & 64 & 128
            & 16 & 32 & 64
            & 8 & 16 & 32 \\
            \midrule
            \textbf{FVD} ($\downarrow$)
            & 145.0 & 147.4 & \underline{135.6} & 142.3
            & 143.6 & \textbf{135.0} & 143.5
            & 151.2 & 149.7 & 146.6 \\
            \textbf{Final loss} ($\downarrow$)
            & 0.0151 & 0.0150 & 0.0147 & 0.0143
            & 0.0146 & 0.0143 & \textbf{0.0141}
            & 0.0146 & 0.0146 & \underline{0.0142} \\
            \midrule
            \textbf{Denoiser Params} (M)
            & 682.2 & 679.2 & 679.7 & 680.8
            & 678.7 & 678.8 & 679.0
            & 681.0 & 683.3 & 688.1 \\
            \textbf{Denoiser FLOPs} (TFLOPs)
            & 1.089 & 1.083 & 1.084 & 1.086
            & 1.082 & 1.082 & 1.082
            & 1.086 & 1.091 & 1.101 \\
            \textbf{Denoiser Latency} (ms, $\downarrow$)
            & 16.07 & 15.96 & 15.99 & 15.58
            & 15.53 & 15.59 & 15.59
            & 15.42 & 15.64 & 15.89 \\
            \textbf{Denoiser Peak} (GiB)
            & 2.674 & 2.650 & 2.652 & 2.659
            & 2.648 & 2.648 & 2.648
            & 2.658 & 2.673 & 2.721 \\
            \midrule
            \textbf{Emb. Params} (M)
            & 3.540 & 0.542 & 1.082 & 2.164
            & 0.087 & 0.173 & 0.345
            & 2.361 & 4.720 & 9.439 \\
            \textbf{Emb. FLOPs} (GFLOPs)
            & 7.248 & 1.107 & 2.215 & 4.429
            & 0.252 & 0.503 & 1.007
            & 4.882 & 9.764 & 19.529 \\
            \textbf{Emb. Latency} (ms, $\downarrow$)
            & 1.095 & 1.047 & 1.059 & 1.072
            & 0.382 & 0.383 & 0.245
            & 0.416 & 0.478 & 0.676 \\
            \textbf{Emb. Peak} (MiB)
            & 89.257 & 51.817 & 56.630 & 68.254
            & 34.669 & 35.333 & 37.536
            & 47.508 & 78.008 & 136.510 \\
            \bottomrule
        \end{tabular}
    }
\end{table}

\noindent\textbf{Register tokens and guidance.}
In Tab.~\ref{tab:register-c2v}, we ablate both the information carried by register tokens and the effect of register guidance~(RG) in class-conditioned generation. Under the default 50-step Euler sampling setup, adding registers without RG yields no FVD improvement over the register-free baseline. With RG, however, all tested register variants outperform this baseline, regardless of the conditioning information they carry. The 50-step Heun and 100-step Euler results further show that RG retains its advantage over higher-NFE baselines using more sampling steps or a higher-order solver, consistent with findings of prior works~\citep{reynaud2024jvid,lu2025dpmsolverplusplus,salajegheh2026higher}.

\begin{wrapfigure}{r}{0.45\textwidth}
    \centering
    \resizebox{!}{0.9\linewidth}{\input{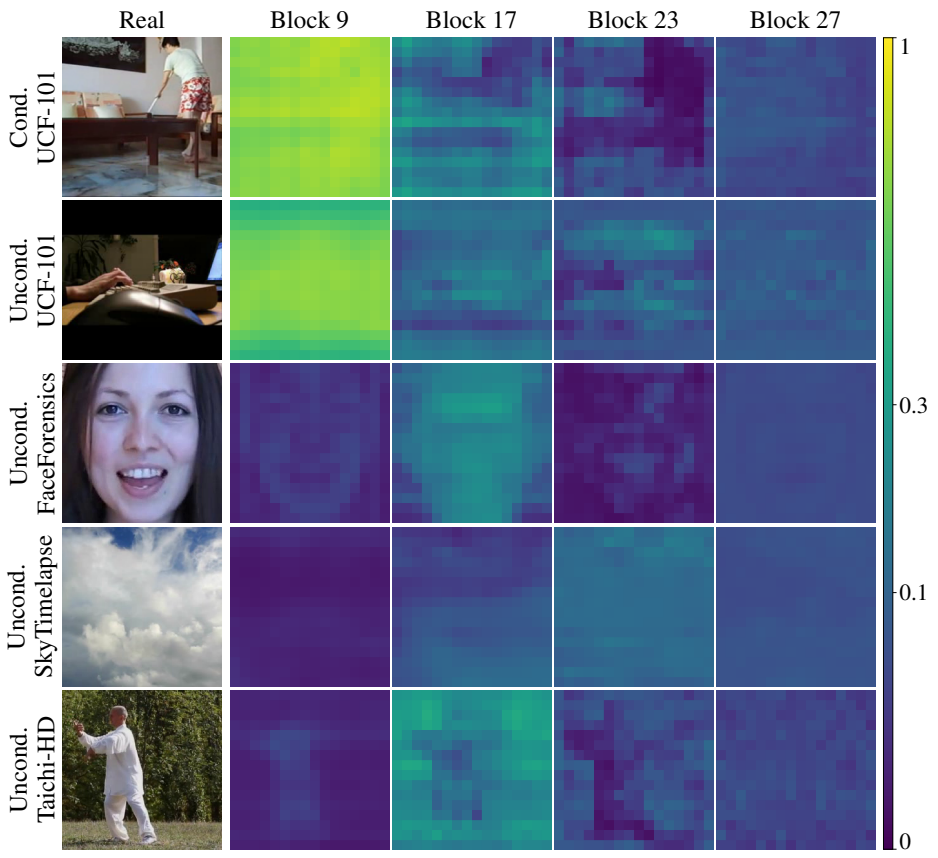}}
    \caption{\textbf{Patch-to-register attention interactions across Transformer blocks.} Register tokens are prepended to visual tokens after the first $L_1=9$ blocks.}
    \label{fig:patch-register-attention}
\end{wrapfigure}

Although prior studies demonstrate the effectiveness of in-context class conditioning for pixel-space image generation~\citep{starodubcev2026registerguidance,li2026jit,lu2026pmf}, we observe no clear benefit from any of the three variants of register-based conditioning in Tab.~\ref{tab:register-c2v}. Instead, registers with $e_r=0$ achieve the lowest FVD of 129.2. These results suggest that, under this sampling setup, the benefits of register tokens depend on RG, rather than register-based conditioning..

\begin{table}[t]
    \centering
    \caption{\textbf{Register and sampling ablations for class-conditioned video generation on UCF-101.} NFE denotes the number of function evaluations, and No reg. denotes the register-free baseline.}
    \label{tab:register-c2v}
    \small
    \setlength{\tabcolsep}{3.5pt}
    \renewcommand{\arraystretch}{1.15}
    \begin{tabular*}{\linewidth}{@{\extracolsep{\fill}}cccccccccc@{}}
        \toprule
        \multicolumn{2}{c}{\textbf{Guidance}}
        & \multicolumn{3}{c}{\textbf{Sampling}}
        & \multicolumn{5}{c}{\textbf{FVD@2048} ($\downarrow$)} \\
        \cmidrule(lr){1-2} \cmidrule(lr){3-5} \cmidrule(lr){6-10}
        CFG & RG & Solver & Steps & NFE
        & No reg. & $e_r=0$ & $e_r=e_y$ & $e_r=e_t$ & $e_r=e_y+e_t$ \\
        \midrule
        \cmark & \xmark & Euler & 50  & 100 &
        \textbf{136.5} & 137.9 & 138.6 & \underline{137.2} & \underline{140.4} \\
        \cmark & \xmark & Euler & 100 & 200 &
        \underline{145.2} & \underline{135.1} & \underline{136.8} & 138.0 & 142.5 \\
        \cmark & \xmark & Heun  & 50  & 200 &
        167.4 & 158.7 & 149.1 & 157.2 & 160.8 \\
        \cmark & \cmark & Euler & 50  & 150 &
        -- & \textbf{129.2} & \textbf{130.4} & \textbf{133.3} & \textbf{132.0} \\
        \bottomrule
    \end{tabular*}
\end{table}

As shown in Tab.~\ref{tab:fvd-main}, RG works well on UCF-101, but brings no obvious gains on the other 3 datasets. Without register guidance, JVT variants with and without registers exhibit broadly similar FVD scores and intermediate feature-map smoothness. However, as illustrated in Fig.~\ref{fig:patch-register-attention}, visual tokens attend strongly to register tokens immediately after their insertion ($L_1=9$) on UCF-101, while this interaction is noticeably weaker on FaceForensics, SkyTimelapse, and Taichi-HD. 

This dataset-dependent difference may be attributed to the global semantics encoded by register tokens~\citep{starodubcev2026registerguidance}. UCF-101 spans 101 action categories and exhibits greater semantic diversity than the other three datasets. We hypothesize that this diversity, together with stronger patch-to-register interactions, makes the register-enabled prediction $v_r$ a more informative guide for its register-free counterpart $v_{nr}$. In contrast, the more homogeneous semantics of the other datasets may limit the benefits of RG.

\begin{wraptable}{r}{0.5\textwidth}
    \centering
    \caption{\textbf{Ablation study of AdaLN on UCF-101.} We report class-conditioned FVD@2048. -- denotes removal of AdaLN or register tokens.}
    \label{tab:adaln-ablation}
    \small
    \begin{tabular*}{\linewidth}{@{}lccc@{}}
        \toprule
        \textbf{Variant} & \textbf{AdaLN condition} & $\boldsymbol{e_r}$ & \textbf{FVD} ($\downarrow$) \\
        \midrule
        baseline     & $e_t+e_y$ & --    & \textbf{136.5} \\
        w/ $e_r=e_y$ & $e_t+e_y$ & $e_y$ & \underline{138.6} \\
        w/o noise level & $e_y$     & --    & 140.6 \\
        w/o AdaLN    & -- & $e_y$ & 156.3 \\
        \bottomrule
    \end{tabular*}
\end{wraptable}

\noindent\textbf{Adaptive layer normalization.}
Standard DiTs inject noise level and class information through adaptive layer normalization~(AdaLN)~\citep{peebles2023dit}. However, recent works start to explore AdaLN-free DiTs for image generation, using in-context tokens, cross-attention, or multimodal joint attention for conditioning~\citep{geng2026imf,lu2026pmf,zeng2026i1,wang2026minit2i}. In Tab.~\ref{tab:adaln-ablation}, we examine whether register-based conditioning offers a reliable alternative to AdaLN in JVT. Relative to the baseline, neither adding class-conditioned registers nor removing noise level conditioning improves FVD. With class-only conditioning, replacing AdaLN with register tokens further worsens FVD to 156.3. We also observe no clear efficiency gain from removing AdaLN. Unconditional ablations show no consistent FVD improvement from jointly removing noise level conditioning and AdaLN (Appendix~\ref{sec:appendix_ablations}, Tab.~\ref{tab:adaln-ablation-uncond}). These results support retaining standard AdaLN as JVT's default conditioning mechanism.

\subsection{Camera-conditioned Video Generation}\label{sec:camera-conditioned-generation}

\begin{figure}[t]
    \centering
    \includegraphics[width=\linewidth]{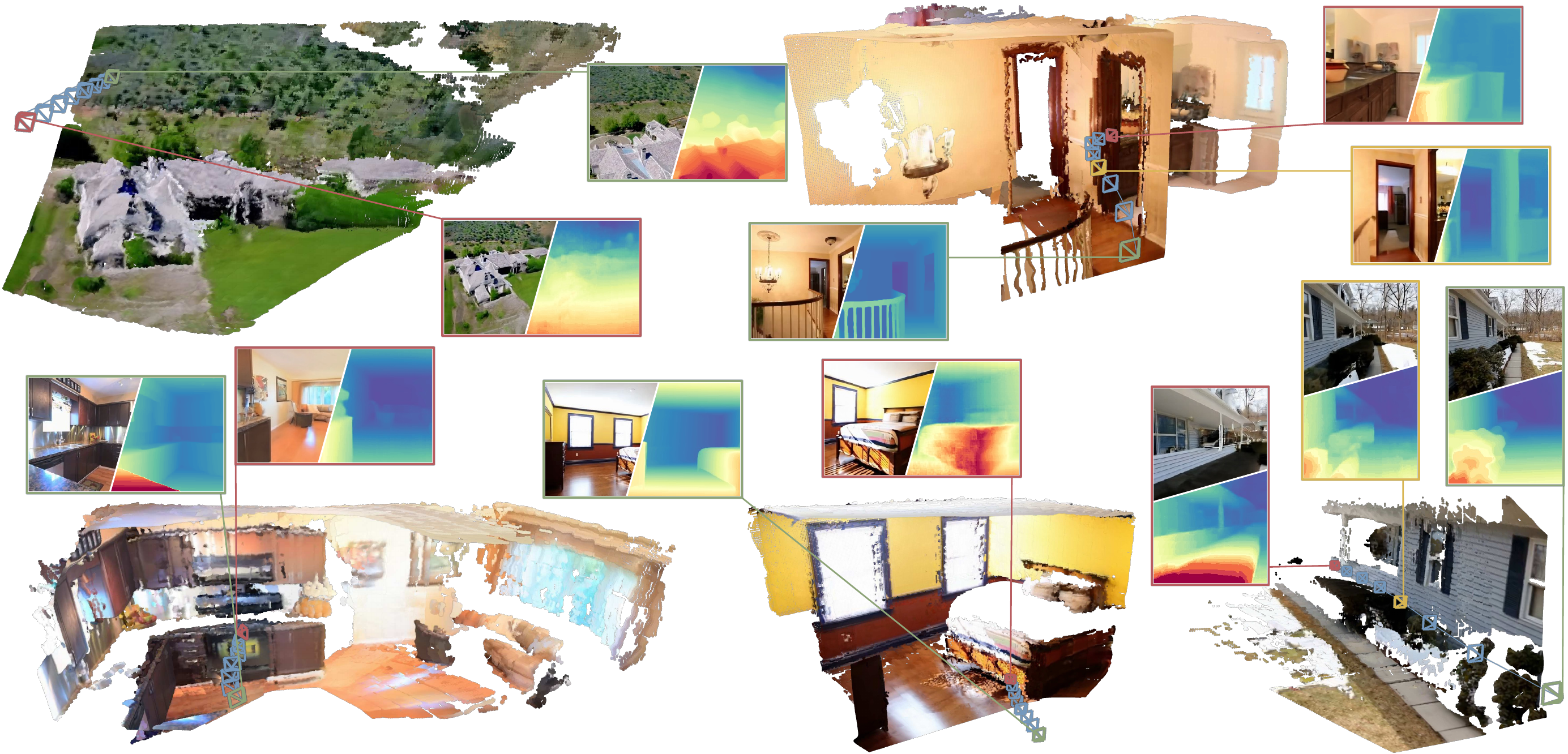}
    \caption{\textbf{Joint RGB-disparity generation and point-cloud reconstruction on RealEstate10K}. We show generated RGB frames alongside aligned disparity maps. Colored point clouds are obtained by unprojecting depth recovered from disparity using the camera parameters, without an additional reconstruction model at inference.}
    \label{fig:joint-pcd}
\end{figure}

We instantiate the pixel-aligned joint generation framework in Sec.~\ref{sec:method_joint_generation} for camera-conditioned video generation. Given an initial RGB frame and a target camera trajectory, we jointly generate RGB frames and aligned disparity maps. After recovering depth from disparity, the camera parameters allow the predictions to be unprojected into a sequence of colored point clouds, providing an explicit 4D representation of the generated scene, as shown in Fig.~\ref{fig:joint-pcd}.

\noindent\textbf{Experimental setting.}
We conduct 8-frame camera-conditioned video generation experiments on RealEstate10K~\citep{zhou2018realestate10k} at a resolution of $256\times256$. As dense depth annotations are unavailable, a frozen lightweight 3D reconstruction model, Pi3X~\citep{wang2026pi3} supplies depth targets online during training. Following prior works~\citep{song2026depthmaster,zhu2025aether}, we reversibly transform depth $D$ to square-root disparity $d=(D+1)^{-1/2}$ and concatenate it with RGB as a four-channel target.

\noindent\textbf{Evaluation.}
We evaluate RGB video quality using FVD, and assess camera following by comparing Pi3X-estimated relative poses with the target camera trajectory. Following prior works~\citep{wang2024dust3r,wang2025vggt,wang2026pi3}, we report relative rotation accuracy (RRA), relative translation accuracy (RTA), and AUC as camera-following metrics. Specifically, RRA and RTA measure the percentages of evaluated frame pairs whose rotation and translation-direction errors fall below a given threshold, respectively, while AUC summarizes their joint error distribution. We use thresholds of $5^\circ$ and $30^\circ$.

\begin{table}[htp]
    \centering
    \caption{\textbf{Camera-conditioned video generation on RealEstate10K.} FVD evaluates RGB video quality. Relative rotation accuracy (RRA), relative translation accuracy (RTA), and AUC evaluate agreement between the generated video and the target camera trajectory.}
    \label{tab:camera-conditioned-joint-prediction}
    \small
    \setlength{\tabcolsep}{3.5pt}
    \resizebox{\linewidth}{!}{
        \begin{tabular}{@{}lcccccccc@{}}
            \toprule
            \multirow{2}{*}{\textbf{Method}} &
            \multirow{2}{*}{\textbf{Objective}} &
            \multirow{2}{*}{\textbf{FVD} ($\downarrow$)} &
            \multicolumn{2}{c}{\textbf{RRA} ($\uparrow$)} &
            \multicolumn{2}{c}{\textbf{RTA} ($\uparrow$)} &
            \multicolumn{2}{c}{\textbf{AUC} ($\uparrow$)} \\
            \cmidrule(lr){4-5} \cmidrule(lr){6-7} \cmidrule(lr){8-9}
            & & &
            \textbf{@5} & \textbf{@30} &
            \textbf{@5} & \textbf{@30} &
            \textbf{@5} & \textbf{@30} \\
            \midrule
            DFoT~\citep{song2025historyguided} & RGB & 35.3 & \textbf{99.9} & 100 & \underline{77.7} & \underline{97.4} & \underline{58.3} & \underline{87.2} \\
            Geometry Forcing~\citep{wu2026geometryforcing} & RGB & 46.8 & \underline{99.8} & 100 & 73.6 & 96.4 & 52.1 & 84.7 \\
            LIVE~\citep{huang2026live} & RGB & 42.5 & 99.1 & 100 & 49.8 & 92.6 & 28.4 & 73.3 \\
            JVT~(Ours) & RGB & \textbf{27.0} & \textbf{99.9} & 100 & 72.4 & 96.0 & 52.8 & 84.1 \\
            JVT~(Ours) & RGB-disparity & \underline{27.5} & \textbf{99.9} & 100 & \textbf{79.5} & \textbf{98.4} & \textbf{58.6} & \textbf{88.6} \\
            \bottomrule
        \end{tabular}
    }
\end{table}

\begin{figure}[t]
    \centering
    \resizebox{\linewidth}{!}{\input{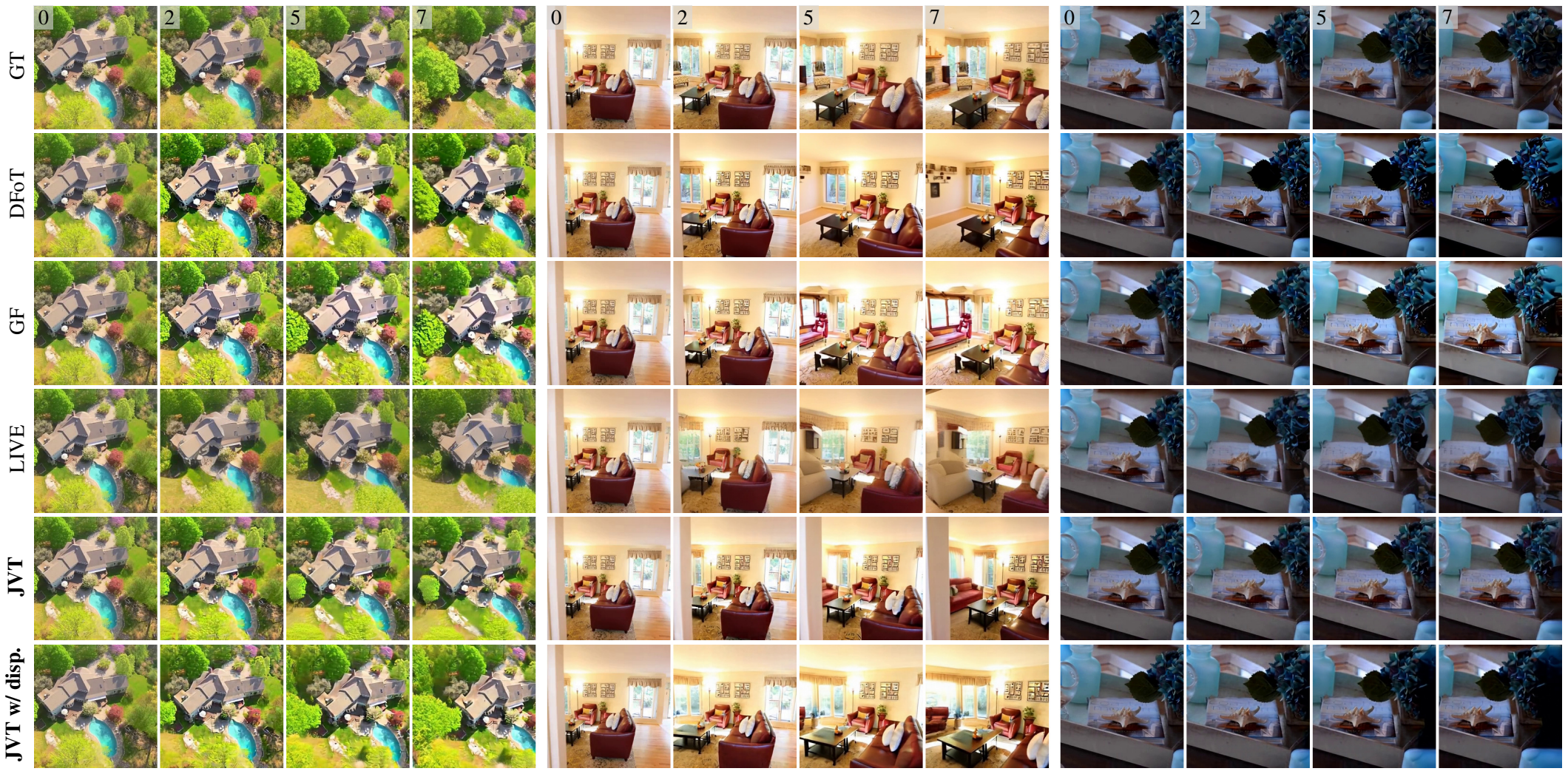}}
    \caption{\textbf{Qualitative comparison of camera-conditioned video generation on RealEstate10K}, with four selected frames per scene. GT denotes ground truth and GF denotes Geometry Forcing. JVT and JVT w/ disp. denote the RGB-only and joint RGB-disparity variants, respectively.}
    \label{fig:re10k-three-scenes}
\end{figure}

We compare JVT with pixel-space U-ViT models~\citep{song2025historyguided,wu2026geometryforcing} and a latent-space DiT~\citep{huang2026live}. As shown in Tab.~\ref{tab:camera-conditioned-joint-prediction}, both JVT variants achieve lower FVD than these baselines. Joint RGB-disparity generation achieves the highest RTA and AUC among all compared methods, with only a slight increase in FVD relative to RGB-only JVT. The qualitative results in Fig.~\ref{fig:re10k-three-scenes} further illustrate the differences in camera following.

Across methods, RRA is consistently high, whereas RTA remains substantially lower. This gap highlights a key limitation of these camera-conditioned video generators: they can follow prescribed camera rotations relatively reliably, but are less accurate at reproducing camera translations. Joint RGB-disparity generation narrows this gap, suggesting that explicit depth cues help the model relate camera motion to scene geometry. Within a unified pixel-space generative framework, JVT natively supports this joint modeling of RGB and geometry and improves camera following.


\section{Conclusion}

We demonstrate pixel-space video generation with a simple DiT-style Transformer, achieving generation quality competitive with latent diffusion baselines in both class-conditioned and unconditional video generation benchmarks. Our ablations study suggests patch-first embedding for spatiotemporal tokenization and standard AdaLN for conditioning. For the register mechanism, register-free JVT provides an acceptable baseline, while register guidance can offer additional gains, particularly on the more diverse UCF-101 dataset.

Pixel-space modeling makes pixel-aligned joint generation straightforward by representing RGB and aligned modalities on a common spatiotemporal grid, without encoding them into a shared pretrained latent space. When extending JVT to camera-conditioned joint RGB-disparity generation, we find camera following is improved relative to RGB-only baselines, with a small FVD trade-off.

These findings highlight the potential of simple DiT-style Transformers for pixel-space video generation and pixel-aligned joint modeling. We hope our empirical findings help guide future efforts to scale pixel-space video generation to larger model size, more datasets, longer videos and higher resolutions.

\bibliography{jvt}

@inproceedings{leng2025repae,
    author    = {Leng, Xingjian and Singh, Jaskirat and Hou, Yunzhong and Xing, Zhenchang and Xie, Saining and Zheng, Liang},
    title     = {REPA-E: Unlocking VAE for End-to-End Tuning of Latent Diffusion Transformers},
    booktitle = {Proceedings of the IEEE/CVF International Conference on Computer Vision (ICCV)},
    month     = {October},
    year      = {2025},
    pages     = {18262-18272}
}

@inproceedings{cheng2025leanvae,
    author    = {Cheng, Yu and Yuan, Fajie},
    title     = {LeanVAE: An Ultra-Efficient Reconstruction VAE for Video Diffusion Models},
    booktitle = {Proceedings of the IEEE/CVF International Conference on Computer Vision (ICCV)},
    month     = {October},
    year      = {2025},
    pages     = {15692-15702}
}

@inproceedings{wang2026eliminating,
    title={Eliminating {VAE} for Fast and High-Resolution Generative Detail Restoration},
    author={Yan Wang and Shijie Zhao and Junlin Li and Li zhang},
    booktitle={The Fourteenth International Conference on Learning Representations},
    year={2026},
    url={https://openreview.net/forum?id=1uu4Hg2Nlk}
}

@article{bounoua2024mldm,
    title={Multi-modal latent diffusion},
    author={Bounoua, Mustapha and Franzese, Giulio and Michiardi, Pietro},
    journal={Entropy},
    volume={26},
    number={4},
    pages={320},
    year={2024},
    publisher={MDPI}
}

@inproceedings{qiu2025mvae,
    title={Multimodal variational autoencoder: A barycentric view},
    author={Qiu, Peijie and Zhu, Wenhui and Kumar, Sayantan and Chen, Xiwen and Yang, Jin and Sun, Xiaotong and Razi, Abolfazl and Wang, Yalin and Sotiras, Aristeidis},
    booktitle={Proceedings of the AAAI Conference on Artificial Intelligence},
    volume={39},
    number={19},
    pages={20060--20068},
    year={2025}
}

@inproceedings{tong2022videomae,
    author = {Tong, Zhan and Song, Yibing and Wang, Jue and Wang, Limin},
    booktitle = {Advances in Neural Information Processing Systems},
    editor = {S. Koyejo and S. Mohamed and A. Agarwal and D. Belgrave and K. Cho and A. Oh},
    pages = {10078--10093},
    publisher = {Curran Associates, Inc.},
    title = {VideoMAE: Masked Autoencoders are Data-Efficient Learners for Self-Supervised Video Pre-Training},
    url = {https://proceedings.neurips.cc/paper_files/paper/2022/file/416f9cb3276121c42eebb86352a4354a-Paper-Conference.pdf},
    volume = {35},
    year = {2022}
}

@inproceedings{wang2023videomaev2,
    author    = {Wang, Limin and Huang, Bingkun and Zhao, Zhiyu and Tong, Zhan and He, Yinan and Wang, Yi and Wang, Yali and Qiao, Yu},
    title     = {VideoMAE V2: Scaling Video Masked Autoencoders With Dual Masking},
    booktitle = {Proceedings of the IEEE/CVF Conference on Computer Vision and Pattern Recognition (CVPR)},
    month     = {June},
    year      = {2023},
    pages     = {14549-14560}
}

@misc{bardes2024vjepa,
    title={V-{JEPA}: Latent Video Prediction for Visual Representation Learning},
    author={Adrien Bardes and Quentin Garrido and Jean Ponce and Xinlei Chen and Michael Rabbat and Yann LeCun and Mido Assran and Nicolas Ballas},
    year={2024},
    url={https://openreview.net/forum?id=WFYbBOEOtv}
}

@article{assran2025vjepa2,
    title={V-JEPA~2: Self-Supervised Video Models Enable Understanding, Prediction and Planning},
    author={Assran, Mahmoud and Bardes, Adrien and Fan, David and Garrido, Quentin and Howes, Russell and
    Komeili, Mojtaba and Muckley, Matthew and Rizvi, Ammar and Roberts, Claire and Sinha, Koustuv and Zholus, Artem and
    Arnaud, Sergio and Gejji, Abha and Martin, Ada and Robert Hogan, Francois and Dugas, Daniel and
    Bojanowski, Piotr and Khalidov, Vasil and Labatut, Patrick and Massa, Francisco and Szafraniec, Marc and
    Krishnakumar, Kapil and Li, Yong and Ma, Xiaodong and Chandar, Sarath and Meier, Franziska and LeCun, Yann and
    Rabbat, Michael and Ballas, Nicolas},
    journal={arXiv preprint arXiv:2506.09985},
    year={2025}
}

@inproceedings{arnab2021vivit,
    author    = {Arnab, Anurag and Dehghani, Mostafa and Heigold, Georg and Sun, Chen and Lu\v{c}i\'c, Mario and Schmid, Cordelia},
    title     = {ViViT: A Video Vision Transformer},
    booktitle = {Proceedings of the IEEE/CVF International Conference on Computer Vision (ICCV)},
    month     = {October},
    year      = {2021},
    pages     = {6836-6846}
}

@article{howard2017mobilenets0,
    title   = {MobileNets: Efficient Convolutional Neural Networks for Mobile Vision Applications},
    author  = {Andrew G. Howard and Menglong Zhu and Bo Chen and Dmitry Kalenichenko and Weijun Wang and Tobias Weyand and Marco Andreetto and Hartwig Adam},
    year    = {2017},
    journal = {arXiv preprint arXiv: 1704.04861}
}

@inproceedings{chollet2017xception,
    author    = {Chollet, Francois},
    title     = {Xception: Deep Learning With Depthwise Separable Convolutions},
    booktitle = {Proceedings of the IEEE Conference on Computer Vision and Pattern Recognition (CVPR)},
    month     = {July},
    year      = {2017}
}

@inproceedings{li2026jit,
    author    = {Li, Tianhong and He, Kaiming},
    title     = {Back to Basics: Let Denoising Generative Models Denoise},
    booktitle = {Proceedings of the IEEE/CVF Conference on Computer Vision and Pattern Recognition (CVPR)},
    month     = {June},
    year      = {2026},
    pages     = {36115-36125}
}

@article{tishby2000ib,
    title={The information bottleneck method},
    author={Tishby, Naftali and Pereira, Fernando C and Bialek, William},
    journal={arXiv preprint physics/0004057},
    year={2000}
}

@inproceedings{rifai2011contractiveae,
    author =    {Salah Rifai and Pascal Vincent and Xavier Muller and Xavier Glorot and Yoshua Bengio},
    title =     {Contractive Auto-Encoders: Explicit Invariance During Feature Extraction},
    booktitle = {Proceedings of the 28th International Conference on Machine Learning},
    year =      {2011},
}

@article{makhzani2013kae,
    title={K-sparse autoencoders},
    author={Makhzani, Alireza and Frey, Brendan},
    journal={arXiv preprint arXiv:1312.5663},
    year={2013}
}

@inproceedings{alemi2017deepvib,
    title={Deep Variational Information Bottleneck},
    author={Alexander A. Alemi and Ian Fischer and Joshua V. Dillon and Kevin Murphy},
    booktitle={International Conference on Learning Representations},
    year={2017},
    url={https://openreview.net/forum?id=HyxQzBceg}
}

@inproceedings{darcet2024registers,
    title={Vision Transformers Need Registers},
    author={Timoth{\'e}e Darcet and Maxime Oquab and Julien Mairal and Piotr Bojanowski},
    booktitle={The Twelfth International Conference on Learning Representations},
    year={2024},
    url={https://openreview.net/forum?id=2dnO3LLiJ1}
}

@article{starodubcev2026registerguidance,
    title   = {Registers Matter for Pixel-Space Diffusion Transformers},
    author  = {Nikita Starodubcev and Ilia Sudakov and Ilya Drobyshevskiy and Artem Babenko and Dmitry Baranchuk},
    year    = {2026},
    journal = {arXiv preprint arXiv: 2605.16147}
}

@inproceedings{geng2026imf,
    author    = {Geng, Zhengyang and Lu, Yiyang and Wu, Zongze and Shechtman, Eli and Kolter, J. Zico and He, Kaiming},
    title     = {Improved Mean Flows: On the Challenges of Fastforward Generative Models},
    booktitle = {Proceedings of the IEEE/CVF Conference on Computer Vision and Pattern Recognition (CVPR)},
    month     = {June},
    year      = {2026},
    pages     = {30467-30476}
}

@inproceedings{lu2026pmf,
    title={One-step Latent-free Image Generation with Pixel Mean Flows},
    author={Yiyang Lu and Susie Lu and Qiao Sun and Hanhong Zhao and Zhicheng Jiang and Xianbang Wang and Tianhong Li and Zhengyang Geng and Kaiming He},
    booktitle={Forty-third International Conference on Machine Learning},
    year={2026},
    url={https://openreview.net/forum?id=WUK8JIeetF}
}

@article{zeng2026i1,
    title   = {i1: A Simple and Fully Open Recipe for Strong Text-to-Image Models},
    author  = {Zeng, Boya and Luo, Tianze and Pu, Shu and Shen, Jucheng and Lu, Taiming and Sarch, Gabriel and Liu, Zhuang},
    journal = {arXiv preprint arXiv:2606.11289},
    year    = {2026}
}

@misc{wang2026minit2i,
    title  = {MiniT2I: A Minimalist Baseline for Text-to-Image Generation},
    author = {Wang, Xianbang and Zhao, Hanhong and Lu, Yiyang and Zhou, Kangyang and Ma, Linrui and He, Kaiming},
    year   = {2026},
    url    = {https://peppaking8.github.io/#/post/minit2i}
}

@article{jiang2026empirical,
    title   = {An Empirical Study of Training Pixel-Space Text-to-Image Diffusion Models},
    author  = {Dengyang Jiang and Ruoyi Du and Zhennan Chen and Dongyang Liu and Zanyi Wang and Mingzhe Zheng and Xiangpeng Yang and Huanqia Cai and Aiming Hao and Yuming Jiang and Peng Gao and Harry Yang and Steven Hoi},
    year    = {2026},
    journal = {arXiv preprint arXiv: 2608.16887}
}

@inproceedings{ronneberger2015unet,
    author="Ronneberger, Olaf
    and Fischer, Philipp
    and Brox, Thomas",
    editor="Navab, Nassir
    and Hornegger, Joachim
    and Wells, William M.
    and Frangi, Alejandro F.",
    title="U-Net: Convolutional Networks for Biomedical Image Segmentation",
    booktitle="Medical Image Computing and Computer-Assisted Intervention -- MICCAI 2015",
    year="2015",
    publisher="Springer International Publishing",
    address="Cham",
    pages="234--241",
    isbn="978-3-319-24574-4"
}

@inproceedings{dhariwal2021adt,
    author = {Dhariwal, Prafulla and Nichol, Alexander},
    booktitle = {Advances in Neural Information Processing Systems},
    editor = {M. Ranzato and A. Beygelzimer and Y. Dauphin and P.S. Liang and J. Wortman Vaughan},
    pages = {8780--8794},
    publisher = {Curran Associates, Inc.},
    title = {Diffusion Models Beat GANs on Image Synthesis},
    url = {https://proceedings.neurips.cc/paper_files/paper/2021/file/49ad23d1ec9fa4bd8d77d02681df5cfa-Paper.pdf},
    volume = {34},
    year = {2021}
}

@inproceedings{peebles2023dit,
    author    = {Peebles, William and Xie, Saining},
    title     = {Scalable Diffusion Models with Transformers},
    booktitle = {Proceedings of the IEEE/CVF International Conference on Computer Vision (ICCV)},
    month     = {October},
    year      = {2023},
    pages     = {4195-4205}
}

@article{guo2026pixelu,
    title   = {PixelU: A U-Shaped Transformer for Efficient End-to-End Pixel Diffusion},
    author  = {Zipeng Guo and Lichen Ma and Yu He and Xiaolong Fu and Jingling Fu and Junshi Huang and Yan Li},
    year    = {2026},
    journal = {arXiv preprint arXiv: 2606.27760}
}

@inproceedings{ho2020ddpm,
    author = {Ho, Jonathan and Jain, Ajay and Abbeel, Pieter},
    booktitle = {Advances in Neural Information Processing Systems},
    editor = {H. Larochelle and M. Ranzato and R. Hadsell and M.F. Balcan and H. Lin},
    pages = {6840--6851},
    publisher = {Curran Associates, Inc.},
    title = {Denoising Diffusion Probabilistic Models},
    url = {https://proceedings.neurips.cc/paper_files/paper/2020/file/4c5bcfec8584af0d967f1ab10179ca4b-Paper.pdf},
    volume = {33},
    year = {2020}
}

@inproceedings{song2021sgm,
    title={Score-Based Generative Modeling through Stochastic Differential Equations},
    author={Yang Song and Jascha Sohl-Dickstein and Diederik P Kingma and Abhishek Kumar and Stefano Ermon and Ben Poole},
    booktitle={International Conference on Learning Representations},
    year={2021},
    url={https://openreview.net/forum?id=PxTIG12RRHS}
}

@inproceedings{ho2021cfg,
    title={Classifier-Free Diffusion Guidance},
    author={Jonathan Ho and Tim Salimans},
    booktitle={NeurIPS 2021 Workshop on Deep Generative Models and Downstream Applications},
    year={2021},
    url={https://openreview.net/forum?id=qw8AKxfYbI}
}

@inproceedings{karras2022edm,
    author = {Karras, Tero and Aittala, Miika and Aila, Timo and Laine, Samuli},
    booktitle = {Advances in Neural Information Processing Systems},
    doi = {10.52202/068431-1926},
    editor = {S. Koyejo and S. Mohamed and A. Agarwal and D. Belgrave and K. Cho and A. Oh},
    pages = {26565--26577},
    publisher = {Curran Associates, Inc.},
    title = {Elucidating the Design Space of Diffusion-Based Generative Models},
    url = {https://proceedings.neurips.cc/paper_files/paper/2022/file/a98846e9d9cc01cfb87eb694d946ce6b-Paper-Conference.pdf},
    volume = {35},
    year = {2022}
}

@inproceedings{salimans2022vpred,
    title={Progressive Distillation for Fast Sampling of Diffusion Models},
    author={Tim Salimans and Jonathan Ho},
    booktitle={International Conference on Learning Representations},
    year={2022},
    url={https://openreview.net/forum?id=TIdIXIpzhoI}
}

@inproceedings{rombach2022ldm,
    author    = {Rombach, Robin and Blattmann, Andreas and Lorenz, Dominik and Esser, Patrick and Ommer, Bj\"orn},
    title     = {High-Resolution Image Synthesis With Latent Diffusion Models},
    booktitle = {Proceedings of the IEEE/CVF Conference on Computer Vision and Pattern Recognition (CVPR)},
    month     = {June},
    year      = {2022},
    pages     = {10684-10695}
}

@inproceedings{lipman2023flowmatching,
    title={Flow Matching for Generative Modeling},
    author={Yaron Lipman and Ricky T. Q. Chen and Heli Ben-Hamu and Maximilian Nickel and Matthew Le},
    booktitle={The Eleventh International Conference on Learning Representations },
    year={2023},
    url={https://openreview.net/forum?id=PqvMRDCJT9t}
}

@inproceedings{liu2023rectifiedflow,
    title={Flow Straight and Fast: Learning to Generate and Transfer Data with Rectified Flow},
    author={Xingchao Liu and Chengyue Gong and qiang liu},
    booktitle={The Eleventh International Conference on Learning Representations },
    year={2023},
    url={https://openreview.net/forum?id=XVjTT1nw5z}
}

@inproceedings{esser2024scaling,
    title={Scaling Rectified Flow Transformers for High-Resolution Image Synthesis},
    author={Patrick Esser and Sumith Kulal and Andreas Blattmann and Rahim Entezari and Jonas M{\"u}ller and Harry Saini and Yam Levi and Dominik Lorenz and Axel Sauer and Frederic Boesel and Dustin Podell and Tim Dockhorn and Zion English and Robin Rombach},
    booktitle={Forty-first International Conference on Machine Learning},
    year={2024},
    url={https://openreview.net/forum?id=FPnUhsQJ5B}
}

@inproceedings{zheng2026rae,
    title={Diffusion Transformers with Representation Autoencoders},
    author={Boyang Zheng and Nanye Ma and Shengbang Tong and Saining Xie},
    booktitle={The Fourteenth International Conference on Learning Representations},
    year={2026},
    url={https://openreview.net/forum?id=0u1LigJaab}
}

@inproceedings{tulyakov2018mocogan,
    author = {Tulyakov, Sergey and Liu, Ming-Yu and Yang, Xiaodong and Kautz, Jan},
    title = {MoCoGAN: Decomposing Motion and Content for Video Generation},
    booktitle = {Proceedings of the IEEE Conference on Computer Vision and Pattern Recognition (CVPR)},
    month = {June},
    year = {2018}
}

@inproceedings{tian2021mocoganhd,
    title={A Good Image Generator Is What You Need for High-Resolution Video Synthesis},
    author={Yu Tian and Jian Ren and Menglei Chai and Kyle Olszewski and Xi Peng and Dimitris N. Metaxas and Sergey Tulyakov},
    booktitle={International Conference on Learning Representations},
    year={2021},
    url={https://openreview.net/forum?id=6puCSjH3hwA}
}

@article{yan2021videogpt,
    title   = {VideoGPT: Video Generation using VQ-VAE and Transformers},
    author  = {Wilson Yan and Yunzhi Zhang and Pieter Abbeel and Aravind Srinivas},
    year    = {2021},
    journal = {arXiv preprint arXiv: 2104.10157}
}

@inproceedings{yu2022digan,
    title={Generating Videos with Dynamics-aware Implicit Generative Adversarial Networks},
    author={Sihyun Yu and Jihoon Tack and Sangwoo Mo and Hyunsu Kim and Junho Kim and Jung-Woo Ha and Jinwoo Shin},
    booktitle={International Conference on Learning Representations},
    year={2022},
    url={https://openreview.net/forum?id=Czsdv-S4-w9}
}

@inproceedings{skorokhodov2022styleganv,
    author    = {Skorokhodov, Ivan and Tulyakov, Sergey and Elhoseiny, Mohamed},
    title     = {StyleGAN-V: A Continuous Video Generator With the Price, Image Quality and Perks of StyleGAN2},
    booktitle = {Proceedings of the IEEE/CVF Conference on Computer Vision and Pattern Recognition (CVPR)},
    month     = {June},
    year      = {2022},
    pages     = {3626-3636}
}

@inproceedings{hong2023cogvideo,
    title={CogVideo: Large-scale Pretraining for Text-to-Video Generation via Transformers},
    author={Wenyi Hong and Ming Ding and Wendi Zheng and Xinghan Liu and Jie Tang},
    booktitle={International Conference on Learning Representations},
    year={2023},
    url={https://openreview.net/forum?id=rB6TpjAuSRy}
}

@article{he2022lvdm,
    title={Latent video diffusion models for high-fidelity long video generation},
    author={He, Yingqing and Yang, Tianyu and Zhang, Yong and Shan, Ying and Chen, Qifeng},
    journal={arXiv preprint arXiv:2211.13221},
    year={2022}
}

@InProceedings{shen2023mostganv,
    author    = {Shen, Xiaoqian and Li, Xiang and Elhoseiny, Mohamed},
    title     = {MoStGAN-V: Video Generation With Temporal Motion Styles},
    booktitle = {Proceedings of the IEEE/CVF Conference on Computer Vision and Pattern Recognition (CVPR)},
    month     = {June},
    year      = {2023},
    pages     = {5652-5661}
}

@inproceedings{yu2023pvdm,
    author    = {Yu, Sihyun and Sohn, Kihyuk and Kim, Subin and Shin, Jinwoo},
    title     = {Video Probabilistic Diffusion Models in Projected Latent Space},
    booktitle = {Proceedings of the IEEE/CVF Conference on Computer Vision and Pattern Recognition (CVPR)},
    month     = {June},
    year      = {2023},
    pages     = {18456-18466}
}

@inproceedings{mei2023vidm,
    author    = {Kangfu Mei and Vishal M. Patel},
    editor    = {Brian Williams and Yiling Chen and Jennifer Neville},
    title     = {{VIDM:} Video Implicit Diffusion Models},
    booktitle = {Thirty-Seventh {AAAI} Conference on Artificial Intelligence, {AAAI} 2023, Thirty-Fifth Conference on Innovative Applications of Artificial Intelligence, {IAAI} 2023, Thirteenth Symposium on Educational Advances in Artificial Intelligence, {EAAI} 2023, Washington, DC, USA, February 7-14, 2023},
    pages     = {9117-9125},
    publisher = {{AAAI} Press},
    year      = {2023},
    url       = {https://doi.org/10.1609/aaai.v37i8.26094},
    doi       = {10.1609/AAAI.V37I8.26094},
    bibsource = {dblp computer science bibliography, https://dblp.org}
}

@inproceedings{sun2023glober,
    author = {Sun, Mingzhen and Wang, Weining and Qin, Zihan and Sun, Jiahui and Chen, Sihan and Liu, Jing},
    booktitle = {Advances in Neural Information Processing Systems},
    editor = {A. Oh and T. Naumann and A. Globerson and K. Saenko and M. Hardt and S. Levine},
    pages = {76120--76136},
    publisher = {Curran Associates, Inc.},
    title = {GLOBER: Coherent Non-autoregressive Video Generation via GLOBal Guided Video DecodER},
    url = {https://proceedings.neurips.cc/paper_files/paper/2023/file/efe36e55d80a94d1726f660b8d237a0f-Paper-Conference.pdf},
    volume = {36},
    year = {2023}
}

@inproceedings{wang2024omnitokenizer,
    author = {Wang, Junke and Jiang, Yi and Yuan, Zehuan and Peng, Binyue and Wu, Zuxuan and Jiang, Yu-Gang},
    booktitle = {Advances in Neural Information Processing Systems},
    doi = {10.52202/079017-0887},
    editor = {A. Globerson and L. Mackey and D. Belgrave and A. Fan and U. Paquet and J. Tomczak and C. Zhang},
    pages = {28281--28295},
    publisher = {Curran Associates, Inc.},
    title = {OmniTokenizer: A Joint Image-Video Tokenizer for Visual Generation},
    url = {https://proceedings.neurips.cc/paper_files/paper/2024/file/31994923f58ae5b2d661b300bd439107-Paper-Conference.pdf},
    volume = {37},
    year = {2024}
}

@inproceedings{lu2024vdt,
    title={{VDT}: General-purpose Video Diffusion Transformers via Mask Modeling},
    author={Haoyu Lu and Guoxing Yang and Nanyi Fei and Yuqi Huo and Zhiwu Lu and Ping Luo and Mingyu Ding},
    booktitle={International Conference on Learning Representations},
    year={2024},
    url={https://openreview.net/forum?id=Un0rgm9f04}
}

@article{liu2024fvdm,
  title   = {Redefining Temporal Modeling in Video Diffusion: The Vectorized Timestep Approach},
  author  = {Yaofang Liu and Yumeng Ren and Xiaodong Cun and Aitor Artola and Yang Liu and Tieyong Zeng and Raymond H. Chan and Jean-michel Morel},
  year    = {2024},
  journal = {arXiv preprint arXiv: 2410.03160}
}

@InProceedings{zhou2025magi,
    author    = {Zhou, Deyu and Sun, Quan and Peng, Yuang and Yan, Kun and Dong, Runpei and Wang, Duomin and Ge, Zheng and Duan, Nan and Zhang, Xiangyu},
    title     = {Taming Teacher Forcing for Masked Autoregressive Video Generation},
    booktitle = {Proceedings of the IEEE/CVF Conference on Computer Vision and Pattern Recognition (CVPR)},
    month     = {June},
    year      = {2025},
    pages     = {7374-7384}
}

@article{ma2025latte,
    title={Latte: Latent Diffusion Transformer for Video Generation},
    author={Ma, Xin and Wang, Yaohui and Chen, Xinyuan and Jia, Gengyun and Liu, Ziwei and Li, Yuan-Fang and Chen, Cunjian and Qiao, Yu},
    journal={Transactions on Machine Learning Research},
    year={2025}
}

@article{le2026framedit,
    title   = {FrameDiT: Diffusion Transformer with Matrix Attention for Efficient Video Generation},
    author  = {Minh Khoa Le and Kien Do and Duc Thanh Nguyen and Truyen Tran},
    year    = {2026},
    journal = {arXiv preprint arXiv: 2603.09721}
}

@inproceedings{ho2022vdm,
    author = {Ho, Jonathan and Salimans, Tim and Gritsenko, Alexey and Chan, William and Norouzi, Mohammad and Fleet, David},
    booktitle = {Advances in Neural Information Processing Systems},
    editor = {S. Koyejo and S. Mohamed and A. Agarwal and D. Belgrave and K. Cho and A. Oh},
    pages = {8633--8646},
    publisher = {Curran Associates, Inc.},
    title = {Video Diffusion Models},
    url = {https://proceedings.neurips.cc/paper_files/paper/2022/file/39235c56aef13fb05a6adc95eb9d8d66-Paper-Conference.pdf},
    volume = {35},
    year = {2022}
}

@inproceedings{unterthiner2019fvd,
    title={{FVD}: A new Metric for Video Generation},
    author={Thomas Unterthiner and Sjoerd van Steenkiste and Karol Kurach and Rapha{\"e}l Marinier and Marcin Michalski and Sylvain Gelly},
    booktitle={International Conference on Learning Representations},
    year={2019},
    url={https://openreview.net/forum?id=rylgEULtdN}
}

@inproceedings{yu2023magvit,
    author    = {Yu, Lijun and Cheng, Yong and Sohn, Kihyuk and Lezama, Jos\'e and Zhang, Han and Chang, Huiwen and Hauptmann, Alexander G. and Yang, Ming-Hsuan and Hao, Yuan and Essa, Irfan and Jiang, Lu},
    title     = {MAGVIT: Masked Generative Video Transformer},
    booktitle = {Proceedings of the IEEE/CVF Conference on Computer Vision and Pattern Recognition (CVPR)},
    month     = {June},
    year      = {2023},
    pages     = {10459-10469}
}

@inproceedings{yu2024magvitv2,
    author = {Yu, Lijun and Lezama, Jos\'{e} and Gundavarapu, Nitesh Bharadwaj and Versari, Luca and Sohn, Kihyuk and Minnen, David and Cheng, Yong and Gupta, Agrim and Gu, Xiuye and Hauptmann, Alexander G and Gong, Boqing and Yang, Ming-Hsuan and Essa, Irfan and Ross, David and Jiang, Lu},
    booktitle = {International Conference on Learning Representations},
    editor = {B. Kim and Y. Yue and S. Chaudhuri and K. Fragkiadaki and M. Khan and Y. Sun},
    pages = {765--783},
    title = {Language Model Beats Diffusion - Tokenizer is key to visual generation},
    url = {https://proceedings.iclr.cc/paper_files/paper/2024/file/036912a83bdbb1fd792baf6532f102d8-Paper-Conference.pdf},
    volume = {2024},
    year = {2024}
}

@inproceedings{goyal2025magnets,
    title={Masked Generative Nested Transformers with Decode Time Scaling},
    author={Sahil Goyal and Debapriya Tula and Gagan Jain and Pradeep Shenoy and Prateek Jain and Sujoy Paul},
    booktitle={Forty-second International Conference on Machine Learning},
    year={2025},
    url={https://openreview.net/forum?id=h4lZMgGyAz}
}

@inproceedings{wang2025par,
    author    = {Wang, Yuqing and Ren, Shuhuai and Lin, Zhijie and Han, Yujin and Guo, Haoyuan and Yang, Zhenheng and Zou, Difan and Feng, Jiashi and Liu, Xihui},
    title     = {Parallelized Autoregressive Visual Generation},
    booktitle = {Proceedings of the IEEE/CVF Conference on Computer Vision and Pattern Recognition (CVPR)},
    month     = {June},
    year      = {2025},
    pages     = {12955-12965}
}

@article{goyal2026elt,
    title   = {ELT: Elastic Looped Transformers for Visual Generation},
    author  = {Sahil Goyal and Swayam Agrawal and Gautham Govind Anil and Prateek Jain and Sujoy Paul and Aditya Kusupati},
    year    = {2026},
    journal = {arXiv preprint arXiv: 2604.09168}
}

@inproceedings{sun2025ardiffusion,
    author    = {Sun, Mingzhen and Wang, Weining and Li, Gen and Liu, Jiawei and Sun, Jiahui and Feng, Wanquan and Lao, Shanshan and Zhou, Siyu and He, Qian and Liu, Jing},
    title     = {AR-Diffusion: Asynchronous Video Generation with Auto-Regressive Diffusion},
    booktitle = {Proceedings of the IEEE/CVF Conference on Computer Vision and Pattern Recognition (CVPR)},
    month     = {June},
    year      = {2025},
    pages     = {7364-7373}
}

@inproceedings{gao2025ca2vdm,
    title={Ca2-{VDM}: Efficient Autoregressive Video Diffusion Model with Causal Generation and Cache Sharing},
    author={Kaifeng Gao and Jiaxin Shi and Hanwang Zhang and Chunping Wang and Jun Xiao and Long Chen},
    booktitle={Forty-second International Conference on Machine Learning},
    year={2025},
    url={https://openreview.net/forum?id=YbtH1aoE1V}
}

@article{hu2026acdit,
    title={{ACD}iT: Interpolating Autoregressive Conditional Modeling and Diffusion Transformer},
    author={Jinyi Hu and Shengding Hu and Yuxuan Song and Yufei Huang and Mingxuan Wang and Hao Zhou and Zhiyuan Liu and Wei-Ying Ma and Maosong Sun},
    journal={Transactions on Machine Learning Research},
    issn={2835-8856},
    year={2026},
    url={https://openreview.net/forum?id=OuFNXESoCO},
    note={}
}

@article{gu2025far,
  title   = {Long-Context Autoregressive Video Modeling with Next-Frame Prediction},
  author  = {Yuchao Gu and Weijia Mao and Mike Zheng Shou},
  year    = {2025},
  journal = {arXiv preprint arXiv: 2503.19325}
}

@inproceedings{song2025historyguided,
    title={History-Guided Video Diffusion},
    author={Kiwhan Song and Boyuan Chen and Max Simchowitz and Yilun Du and Russ Tedrake and Vincent Sitzmann},
    booktitle={Forty-second International Conference on Machine Learning},
    year={2025},
    url={https://openreview.net/forum?id=j8Vr3E3vhy}
}

@inproceedings{wu2026geometryforcing,
    title={Geometry Forcing: Marrying Video Diffusion and 3D Representation for Consistent World Modeling},
    author={Haoyu Wu and Diankun Wu and Tianyu He and Junliang Guo and Yang Ye and Yueqi Duan and Jiang Bian},
    booktitle={The Fourteenth International Conference on Learning Representations},
    year={2026},
    url={https://openreview.net/forum?id=ULXYZCms41}
}

@inproceedings{huang2026live,
    title={{LIVE}: Long-horizon Interactive Video World Modeling},
    author={Junchao Huang and Ziyang Ye and Xinting Hu and Tianyu He and Guiyu Zhang and Shaoshuai Shi and Jiang Bian and Li Jiang},
    booktitle={Forty-third International Conference on Machine Learning},
    year={2026},
    url={https://openreview.net/forum?id=iKKZmgxkbG}
}

@article{soomro2012ucf101,
    title   = {UCF101: A Dataset of 101 Human Actions Classes From Videos in The Wild},
    author  = {Khurram Soomro and Amir Roshan Zamir and Mubarak Shah},
    year    = {2012},
    journal = {arXiv preprint arXiv: 1212.0402}
}

@article{rossler2018faceforensics,
    title   = {FaceForensics: A Large-scale Video Dataset for Forgery Detection in Human Faces},
    author  = {R{\"o}ssler, Andreas and Cozzolino, Davide and Verdoliva, Luisa and Riess, Christian and Thies, Justus and Nie{\ss}ner, Matthias},
    year    = {2018},
    journal = {arXiv preprint arXiv: 1803.09179}
}

@inproceedings{xiong2018skytimelapse,
    author = {Xiong, Wei and Luo, Wenhan and Ma, Lin and Liu, Wei and Luo, Jiebo},
    title = {Learning to Generate Time-Lapse Videos Using Multi-Stage Dynamic Generative Adversarial Networks},
    booktitle = {Proceedings of the IEEE Conference on Computer Vision and Pattern Recognition (CVPR)},
    month = {June},
    year = {2018}
}

@inproceedings{siarohin2019taichi,
    author = {Siarohin, Aliaksandr and Lathuili\`{e}re, St\'{e}phane and Tulyakov, Sergey and Ricci, Elisa and Sebe, Nicu},
    booktitle = {Advances in Neural Information Processing Systems},
    editor = {H. Wallach and H. Larochelle and A. Beygelzimer and F. d\textquotesingle Alch\'{e}-Buc and E. Fox and R. Garnett},
    pages = {},
    publisher = {Curran Associates, Inc.},
    title = {First Order Motion Model for Image Animation},
    url = {https://proceedings.neurips.cc/paper_files/paper/2019/file/31c0b36aef265d9221af80872ceb62f9-Paper.pdf},
    volume = {32},
    year = {2019}
}

@article{zhou2018realestate10k,
    title={Stereo magnification: Learning view synthesis using multiplane images},
    author={Zhou, Tinghui and Tucker, Richard and Flynn, John and Fyffe, Graham and Snavely, Noah},
    journal={arXiv preprint arXiv:1805.09817},
    year={2018}
}

@article{song2026depthmaster,
    author={Song, Ziyang and Wang, Zerong and Li, Bo and Zhang, Hao and Zhu, Ruijie and Liu, Li and Jiang, Peng-Tao and Zhang, Tianzhu},
    journal={IEEE Transactions on Circuits and Systems for Video Technology}, 
    title={DepthMaster: Taming Diffusion Models for Monocular Depth Estimation}, 
    year={2026},
    volume={},
    number={},
    pages={1-1},
    doi={10.1109/TCSVT.2026.3681436}
}

@inproceedings{zhu2025aether,
    author    = {Zhu, Haoyi and Wang, Yifan and Zhou, Jianjun and Chang, Wenzheng and Zhou, Yang and Li, Zizun and Chen, Junyi and Shen, Chunhua and Pang, Jiangmiao and He, Tong},
    title     = {Aether: Geometric-Aware Unified World Modeling},
    booktitle = {Proceedings of the IEEE/CVF International Conference on Computer Vision (ICCV)},
    month     = {October},
    year      = {2025},
    pages     = {8535-8546}
}

@inproceedings{ke2024marigold,
    author    = {Ke, Bingxin and Obukhov, Anton and Huang, Shengyu and Metzger, Nando and Daudt, Rodrigo Caye and Schindler, Konrad},
    title     = {Repurposing Diffusion-Based Image Generators for Monocular Depth Estimation},
    booktitle = {Proceedings of the IEEE/CVF Conference on Computer Vision and Pattern Recognition (CVPR)},
    month     = {June},
    year      = {2024},
    pages     = {9492-9502}
}

@article{chen2025deepverse,
    title={DeepVerse: 4D Autoregressive Video Generation as a World Model},
    author={Junyi Chen and Haoyi Zhu and Xianglong He and Yifan Wang and Jianjun Zhou and Wenzheng Chang and Yang Zhou and Zizun Li and Zhoujie Fu and Jiangmiao Pang and Tong He},
    journal={arXiv preprint arXiv:2506.01103},
    year={2025}
}

@inproceedings{huang2025unityvideo,
    author    = {Huang, Jiehui and Zhang, Yuechen and He, Xu and Gao, Yuan and Cen, Zhi and Xia, Bin and Zhou, Yan and Tao, Xin and Wan, Pengfei and Jia, Jiaya},
    title     = {UnityVideo: Unified Multi-Modal Multi-Task Learning for Enhancing World-Aware Video Generation},
    booktitle = {Proceedings of the IEEE/CVF Conference on Computer Vision and Pattern Recognition (CVPR)},
    month     = {June},
    year      = {2026},
    pages     = {4471-4481}
}

@inproceedings{wang2026pi3,
    title={$\pi^3$: Permutation-Equivariant Visual Geometry Learning},
    author={Yifan Wang and Jianjun Zhou and Haoyi Zhu and Wenzheng Chang and Yang Zhou and Zizun Li and Junyi Chen and Jiangmiao Pang and Chunhua Shen and Tong He},
    booktitle={The Fourteenth International Conference on Learning Representations},
    year={2026},
    url={https://openreview.net/forum?id=DTQIjngDta}
}

@inproceedings{wang2024dust3r,
    author    = {Wang, Shuzhe and Leroy, Vincent and Cabon, Yohann and Chidlovskii, Boris and Revaud, Jerome},
    title     = {DUSt3R: Geometric 3D Vision Made Easy},
    booktitle = {Proceedings of the IEEE/CVF Conference on Computer Vision and Pattern Recognition (CVPR)},
    month     = {June},
    year      = {2024},
    pages     = {20697-20709}
}

@inproceedings{wang2025vggt,
    author    = {Wang, Jianyuan and Chen, Minghao and Karaev, Nikita and Vedaldi, Andrea and Rupprecht, Christian and Novotny, David},
    title     = {VGGT: Visual Geometry Grounded Transformer},
    booktitle = {Proceedings of the IEEE/CVF Conference on Computer Vision and Pattern Recognition (CVPR)},
    month     = {June},
    year      = {2025},
    pages     = {5294-5306}
}

@article{ho2022cascaded,
    title={Cascaded diffusion models for high fidelity image generation},
    author={Ho, Jonathan and Saharia, Chitwan and Chan, William and Fleet, David J and Norouzi, Mohammad and Salimans, Tim},
    journal={Journal of Machine Learning Research},
    volume={23},
    number={47},
    pages={1--33},
    year={2022}
}

@inproceedings{hoogeboom2023sid,
    title = 	 {simple diffusion: End-to-end diffusion for high resolution images},
    author =     {Hoogeboom, Emiel and Heek, Jonathan and Salimans, Tim},
    booktitle =  {Proceedings of the 40th International Conference on Machine Learning},
    pages = 	 {13213--13232},
    year = 	     {2023},
    editor = 	 {Krause, Andreas and Brunskill, Emma and Cho, Kyunghyun and Engelhardt, Barbara and Sabato, Sivan and Scarlett, Jonathan},
    volume = 	 {202},
    series = 	 {Proceedings of Machine Learning Research},
    month = 	 {23--29 Jul},
    publisher =  {PMLR},
    url = 	     {https://proceedings.mlr.press/v202/hoogeboom23a.html}
}

@inproceedings{crowson2024hdit,
  title = 	  {Scalable High-Resolution Pixel-Space Image Synthesis with Hourglass Diffusion Transformers},
  author =    {Crowson, Katherine and Baumann, Stefan Andreas and Birch, Alex and Abraham, Tanishq Mathew and Kaplan, Daniel Z and Shippole, Enrico},
  booktitle = {Proceedings of the 41st International Conference on Machine Learning},
  pages = 	  {9550--9575},
  year = 	  {2024},
  editor = 	  {Salakhutdinov, Ruslan and Kolter, Zico and Heller, Katherine and Weller, Adrian and Oliver, Nuria and Scarlett, Jonathan and Berkenkamp, Felix},
  volume = 	 {235},
  series = 	 {Proceedings of Machine Learning Research},
  month = 	 {21--27 Jul},
  publisher =    {PMLR},
  url = 	 {https://proceedings.mlr.press/v235/crowson24a.html},
}

@inproceedings{teng2024relay,
    title={Relay Diffusion: Unifying diffusion process across resolutions for image synthesis},
    author={Jiayan Teng and Wendi Zheng and Ming Ding and Wenyi Hong and Jianqiao Wangni and Zhuoyi Yang and Jie Tang},
    booktitle={The Twelfth International Conference on Learning Representations},
    year={2024},
    url={https://openreview.net/forum?id=qTlcbLSm4p}
}

@article{chen2025pixelflow,
    title={Pixelflow: Pixel-space generative models with flow},
    author={Chen, Shoufa and Ge, Chongjian and Zhang, Shilong and Sun, Peize and Luo, Ping},
    journal={arXiv preprint arXiv:2504.07963},
    year={2025}
}

@inproceedings{wang2026pixnerd,
    title={PixNerd: Pixel Neural Field Diffusion},
    author={Shuai Wang and Ziteng Gao and Chenhui Zhu and Weilin Huang and Limin Wang},
    booktitle={The Fourteenth International Conference on Learning Representations},
    year={2026},
    url={https://openreview.net/forum?id=BDnOrExHmt}
}

@inproceedings{yu2026pixeldit,
    author    = {Yu, Yongsheng and Xiong, Wei and Nie, Weili and Sheng, Yichen and Liu, Shiqiu and Luo, Jiebo},
    title     = {PixelDiT: Pixel Diffusion Transformers for Image Generation},
    booktitle = {Proceedings of the IEEE/CVF Conference on Computer Vision and Pattern Recognition (CVPR)},
    month     = {June},
    year      = {2026},
    pages     = {14273-14282}
}

@inproceedings{ma2026deco,
    author    = {Ma, Zehong and Wei, Longhui and Wang, Shuai and Zhang, Shiliang and Tian, Qi},
    title     = {DeCo: Frequency-Decoupled Pixel Diffusion for End-to-End Image Generation},
    booktitle = {Proceedings of the IEEE/CVF Conference on Computer Vision and Pattern Recognition (CVPR)},
    month     = {June},
    year      = {2026},
    pages     = {43600-43610}
}

@inproceedings{chen2026dip,
    author    = {Chen, Zhennan and Zhu, Junwei and Chen, Xu and Zhang, Jiangning and Hu, Xiaobin and Zhao, Hanzhen and Wang, Chengjie and Yang, Jian and Tai, Ying},
    title     = {DiP: Taming Diffusion Models in Pixel Space},
    booktitle = {Proceedings of the IEEE/CVF Conference on Computer Vision and Pattern Recognition (CVPR)},
    month     = {June},
    year      = {2026},
    pages     = {36136-36146}
}

@article{lin2026frepix,
    title={FREPix: Frequency-Heterogeneous Flow Matching for Pixel-Space Image Generation},
    author={Lin, Mingfeng and Chen, Jiakun and Han, Liang and Nie, Liqiang},
    journal={arXiv preprint arXiv:2605.06421},
    year={2026}
}

@article{he2026hyperdit,
    title={Hyperdit: Hyper-connected transformers for high-fidelity pixel-space diffusion},
    author={He, Yu and Ma, Lichen and Guo, Zipeng and Shan, Xinyuan and Fu, Jingling and Chen, Dong and Huang, Junshi and Li, Yan},
    journal={arXiv preprint arXiv:2605.15741},
    year={2026}
}

@article{ma2026frequencybooster,
    title={FrequencyBooster: Full-Frequency Modeling for High-Fidelity Pixel Diffusion},
    author={Ma, Lichen and Guo, Zipeng and He, Yu and Fu, Xiaolong and Liu, Luohang and Fu, Jingling and Huang, Junshi and Li, Yan},
    journal={arXiv preprint arXiv:2605.17759},
    year={2026}
}

@inproceedings{gu2024matryoshka,
    title={Matryoshka Diffusion Models},
    author={Jiatao Gu and Shuangfei Zhai and Yizhe Zhang and Joshua M. Susskind and Navdeep Jaitly},
    booktitle={The Twelfth International Conference on Learning Representations},
    year={2024},
    url={https://openreview.net/forum?id=tOzCcDdH9O}
}

@inproceedings{bartal2024lumiere,
    title     = {Lumiere: A Space-Time Diffusion Model for Video Generation},
    author    = {Bar-Tal, Omer and Chefer, Hila and Tov, Omer and Herrmann, Charles and Paiss, Roni and Zada, Shiran and Ephrat, Ariel and Hur, Junhwa and Liu, Guanghui and Raj, Amit and Li, Yuanzhen and Rubinstein, Michael and Michaeli, Tomer and Wang, Oliver and Sun, Deqing and Dekel, Tali and Mosseri, Inbar},
    booktitle = {SIGGRAPH Asia 2024 Conference Papers},
    year      = {2024},
}

@inproceedings{menapace2024snapvideo,
    author    = {Menapace, Willi and Siarohin, Aliaksandr and Skorokhodov, Ivan and Deyneka, Ekaterina and Chen, Tsai-Shien and Kag, Anil and Fang, Yuwei and Stoliar, Aleksei and Ricci, Elisa and Ren, Jian and Tulyakov, Sergey},
    title     = {Snap Video: Scaled Spatiotemporal Transformers for Text-to-Video Synthesis},
    booktitle = {Proceedings of the IEEE/CVF Conference on Computer Vision and Pattern Recognition (CVPR)},
    month     = {June},
    year      = {2024},
    pages     = {7038-7048}
}

@inproceedings{skorokhodov2024hpd,
    author    = {Skorokhodov, Ivan and Menapace, Willi and Siarohin, Aliaksandr and Tulyakov, Sergey},
    title     = {Hierarchical Patch Diffusion Models for High-Resolution Video Generation},
    booktitle = {Proceedings of the IEEE/CVF Conference on Computer Vision and Pattern Recognition (CVPR)},
    month     = {June},
    year      = {2024},
    pages     = {7569-7579}
}

@inproceedings{hoogeboom2025sid2,
    author        = {Hoogeboom, Emiel and Mensink, Thomas and Heek, Jonathan and Lamerigts, Kay and Gao, Ruiqi and Salimans, Tim},
    title         = {Simpler Diffusion: 1.5 FID on ImageNet512 with Pixel-space Diffusion},
    booktitle     = {Proceedings of the IEEE/CVF Conference on Computer Vision and Pattern Recognition (CVPR)},
    month         = {June},
    year          = {2025},
    pages         = {18062-18071}
}

@article{kong2024hunyuanvideo,
    title   = {HunyuanVideo: A Systematic Framework For Large Video Generative Models},
    author  = {Weijie Kong and Qi Tian and Zijian Zhang and Rox Min and Zuozhuo Dai and Jin Zhou and Jiangfeng Xiong and Xin Li and Bo Wu and Jianwei Zhang and Kathrina Wu and Qin Lin and Junkun Yuan and Yanxin Long and Aladdin Wang and Andong Wang and Changlin Li and Duojun Huang and Fang Yang and Hao Tan and Hongmei Wang and Jacob Song and Jiawang Bai and Jianbing Wu and Jinbao Xue and Joey Wang and Kai Wang and Mengyang Liu and Pengyu Li and Shuai Li and Weiyan Wang and Wenqing Yu and Xinchi Deng and Yang Li and Yi Chen and Yutao Cui and Yuanbo Peng and Zhentao Yu and Zhiyu He and Zhiyong Xu and Zixiang Zhou and Zunnan Xu and Yangyu Tao and Qinglin Lu and Songtao Liu and Dax Zhou and Hongfa Wang and Yong Yang and Di Wang and Yuhong Liu and Jie Jiang and Caesar Zhong},
    year    = {2024},
    journal = {arXiv preprint arXiv: 2412.03603}
}

@inproceedings{yang2025cogvideox,
    title={CogVideoX: Text-to-Video Diffusion Models with An Expert Transformer},
    author={Zhuoyi Yang and Jiayan Teng and Wendi Zheng and Ming Ding and Shiyu Huang and Jiazheng Xu and Yuanming Yang and Wenyi Hong and Xiaohan Zhang and Guanyu Feng and Da Yin and Yuxuan.Zhang and Weihan Wang and Yean Cheng and Bin Xu and Xiaotao Gu and Yuxiao Dong and Jie Tang},
    booktitle={The Thirteenth International Conference on Learning Representations},
    year={2025},
    url={https://openreview.net/forum?id=LQzN6TRFg9}
}

@article{wan2025wan21,
    title   = {Wan: Open and Advanced Large-Scale Video Generative Models},
    author  = {Team Wan and Ang Wang and Baole Ai and Bin Wen and Chaojie Mao and Chen-Wei Xie and Di Chen and Feiwu Yu and Haiming Zhao and Jianxiao Yang and Jianyuan Zeng and Jiayu Wang and Jingfeng Zhang and Jingren Zhou and Jinkai Wang and Jixuan Chen and Kai Zhu and Kang Zhao and Keyu Yan and Lianghua Huang and Mengyang Feng and Ningyi Zhang and Pandeng Li and Pingyu Wu and Ruihang Chu and Ruili Feng and Shiwei Zhang and Siyang Sun and Tao Fang and Tianxing Wang and Tianyi Gui and Tingyu Weng and Tong Shen and Wei Lin and Wei Wang and Wei Wang and Wenmeng Zhou and Wente Wang and Wenting Shen and Wenyuan Yu and Xianzhong Shi and Xiaoming Huang and Xin Xu and Yan Kou and Yangyu Lv and Yifei Li and Yijing Liu and Yiming Wang and Yingya Zhang and Yitong Huang and Yong Li and You Wu and Yu Liu and Yulin Pan and Yun Zheng and Yuntao Hong and Yupeng Shi and Yutong Feng and Zeyinzi Jiang and Zhen Han and Zhi-Fan Wu and Ziyu Liu},
    year    = {2025},
    journal = {arXiv preprint arXiv: 2503.20314}
}

@inproceedings{salajegheh2026higher,
  title={When Higher Order Hurts: Pre-asymptotic Order Collapse in Generative ODE Sampling: A Theory of Discretization--Learning Interaction},
  author={Salajegheh, Farzad and Mudur, Sudhir},
  booktitle={European Conference on Computer Vision},
  pages={151--167},
  year={2026},
  organization={Springer}
}

@article{reynaud2024jvid,
  title={JVID: Joint Video-Image Diffusion for Visual-Quality and Temporal-Consistency in Video Generation},
  author={Reynaud, Hadrien and Baugh, Matthew and Dombrowski, Mischa and Cechnicka, Sarah and Meng, Qingjie and Kainz, Bernhard},
  journal={arXiv preprint arXiv:2409.14149},
  year={2024}
}

@article{lu2025dpmsolverplusplus,
  title={Dpm-solver++: Fast solver for guided sampling of diffusion probabilistic models},
  author={Lu, Cheng and Zhou, Yuhao and Bao, Fan and Chen, Jianfei and Li, Chongxuan and Zhu, Jun},
  journal={Machine Intelligence Research},
  volume={22},
  number={4},
  pages={730--751},
  year={2025},
  publisher={Springer}
}
\bibliographystyle{jvt}

\appendix
\clearpage
\section{Appendix}

\subsection{Implementation Details}
\label{sec:appendix_implementation}

Our implementation follows the pixel-space clean-data prediction design of JiT~\citep{li2026jit}, with video generation and camera conditioning implemented in the DFoT framework~\citep{song2025historyguided}. Tab.~\ref{tab:implementation-config} specifies model and optimization configurations.

\begin{table}[!htbp]
    \centering
    \caption{\textbf{Model and optimization configurations.} The standard-generation column specifies recipes for class-conditioned and unconditional video generation, and the two camera columns summarize the RGB and RGB-disparity configurations. Registers only used for class-conditioned and unconditional video generation on UCF-101. Global batch size is per-rank local batch size multiplied by world size, without gradient accumulation.}
    \label{tab:implementation-config}
    \small
    \setlength{\tabcolsep}{3pt}
    \renewcommand{\arraystretch}{1.06}
    \begin{tabularx}{\linewidth}{@{}l>{\centering\arraybackslash}X>{\centering\arraybackslash}X>{\centering\arraybackslash}X@{}}
        \toprule
        & \textbf{Standard RGB} & \textbf{Camera RGB} & \textbf{Camera RGB--disp.} \\
        \midrule
        Backbone & JVT-XL & JVT-L & JVT-L, joint I/O \\
        Depth / width / heads & 28 / 1,152 / 16 & 24 / 1,024 / 16 & 24 / 1,024 / 16 \\
        Video size & $16\times256\times256$ & $8\times256\times256$ & $8\times256\times256$ \\
        Input/output channels & 3 & 3 & 4 \\
        Spatial patch size & \multicolumn{3}{c}{$16\times16$} \\
        Temporal factor & 4 & 1 & 1 \\
        Visual tokens & 1,024 & 2,048 & 2,048 \\
        Embedding & Patch-first, $b=32$ & - & - \\
        Registers, when used & 32, after block 9 & - & - \\
        Reg. init. $e_r$, when used & $e_r=0$ & - & - \\
        Initialization & \multicolumn{3}{c}{From scratch} \\
        \midrule
        Optimizer & \multicolumn{3}{c}{AdamW} \\
        Adam $\beta$ & \multicolumn{3}{c}{$(0.9,0.99)$} \\
        Learning rate & $10^{-4}$ & $8\times10^{-5}$ & $8\times10^{-5}$ \\
        Warm-up updates & 4,000 & 4,000 & 1,000 \\
        LR after warm-up & \multicolumn{3}{c}{Constant} \\
        Weight decay & \multicolumn{3}{c}{0.01} \\
        EMA decay & \multicolumn{3}{c}{0.9999} \\
        Global batch size & \multicolumn{3}{c}{$256$} \\
        Noise shift $s$ & \multicolumn{3}{c}{4} \\
        Training noise levels & Shared across a clip & Frame-wise independent & Frame-wise independent \\
        Division floor $t_\epsilon$ & \multicolumn{3}{c}{$0.005$} \\
        Total updates & ${\approx}1$M & ${\approx}300$K & ${\approx}300$K \\
        \bottomrule
    \end{tabularx}
\end{table}

\paragraph{Register tokens.}
As shown in Fig.~\ref{fig:patch-register-attention}, the register mechanism brings clear gains on UCF-101 but only limited benefits on the other datasets, so we use register tokens only for class-conditioned and unconditional video generation on UCF-101. The register initialization ablation in Tab.~\ref{tab:register-c2v} further shows no gain from injecting the timestep or class embedding into the register tokens, and we therefore just set $e_r=0$ when registers are used. The base registers $R_0$ are learnable tokens initialized i.i.d.\ from a zero-mean Gaussian with standard deviation $0.02$. All register-equipped variants are sampled with register guidance, whose scale is selected by a linear search with a step size of 0.1.

\begin{figure}[htbp]
    \centering
    \includegraphics[width=\linewidth]{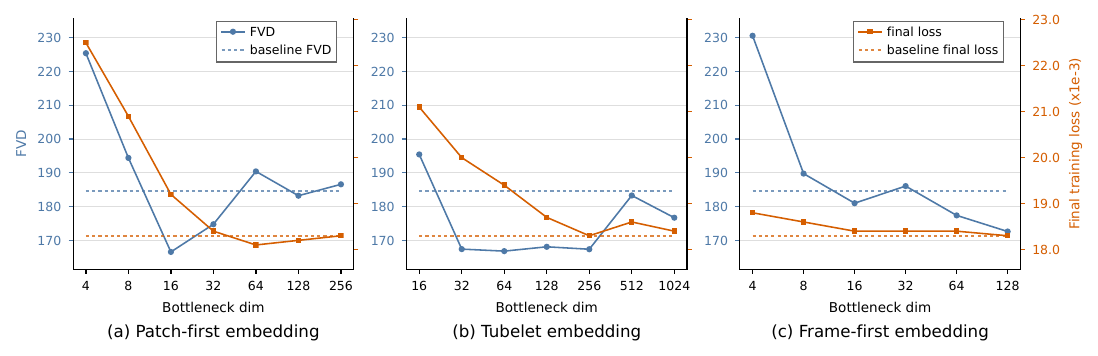}
    \caption{Effect of configured bottleneck width on video patch embedding with batch size 40. FVD uses the left axis and final training loss the right axis. Dashed lines show the corresponding plain-tubelet baseline values. The two vertical scales are independent.}
    \label{fig:fvd-loss-btnk-dim-c2v}
\end{figure}

\noindent\textbf{Training details.}
All models are trained from scratch. Following JiT~\citep{li2026jit}, we use clean data prediction and velocity MSE loss. For noise sampling, we sample noise level $t_{\mathrm{base}}$ from logit-normal distribution~\citep{esser2024scaling} and shift it with noise shift $s$, as shown in Eq.~(\ref{eq:time-shift}). Our full-scale training of JVT use a global batch size of 256 with 8 NVIDIA 80GB GPUs, while some ablations use a batch size of 40 on a single GPU (Fig.~\ref{fig:fvd-loss-btnk-dim-c2v}). 
\begin{equation}
    t
    = \frac{s \cdot t_{\mathrm{base}}}{1 + (s - 1) \cdot t_{\mathrm{base}}}
    \label{eq:time-shift}
\end{equation}
For camera-conditioned generation, we sample the video noise independently for each frame, following DFoT~\citep{song2025historyguided}. We also adopt the data augmentations of the DFoT dataloader: horizontal flipping and temporal reversal, each applied with probability 0.5, and back-and-forth augmentation with probability 0.1.

\noindent\textbf{Sampling configurations.}
Tab.~\ref{tab:implementation-sampling} specifies the exact sampling configuration of each main result. We use the Euler solver for class-conditioned and camera-conditioned generation, and the Heun solver for unconditional generation. All samplers run for 50 denoising steps, with the timestep schedule shifted by the noise shift $s$ (Eq.~(\ref{eq:time-shift})). Depending on the task, we apply classifier-free guidance (CFG)~\citep{ho2021cfg}, register guidance (RG)~\citep{starodubcev2026registerguidance}, or history guidance (HG)~\citep{song2025historyguided}. The CFG scale is selected by a linear search with a step size of 1, and the RG scale with a step size of 0.1. For FVD evaluation, we generate 2,048 clips per method and use the same number of real clips as the reference.

\noindent\textbf{Geometry annotation for RealEstate10K.}
As in Sec.~\ref{sec:camera-conditioned-generation}, dense depth annotations are unavailable for RealEstate10K, so we use a frozen Pi3X~\citep{wang2026pi3} to supply depth pseudo-labels online during training. For each training clip, Pi3X is conditioned on the prescribed camera intrinsics and extrinsics as priors. Since Pi3X predicts scale-invariant geometry and affine-invariant cameras, we estimate the global scale of the predicted depth by aligning the predicted cameras with the ground-truth cameras. The aligned depth is clamped to a minimum of $10^{-5}$ and, following prior works~\citep{song2026depthmaster,zhu2025aether}, reversibly transformed to square-root disparity $d=(D+1)^{-1/2}$, which is concatenated with RGB as a four-channel target.

\begin{table}[!htbp]
    \centering
    \caption{\textbf{Sampling configurations for the main results.} The camera column covers both the RGB and RGB-disparity configurations. CFG, RG, HG denote classifier-free guidance, register guidance and history guidance, respectively. The register guidance scale is selected by a linear search with a step size of 0.1.}
    \label{tab:implementation-sampling}
    \small
    \setlength{\tabcolsep}{3pt}
    \renewcommand{\arraystretch}{1.06}
    \begin{tabularx}{\linewidth}{@{}l>{\centering\arraybackslash}X>{\centering\arraybackslash}X>{\centering\arraybackslash}X>{\centering\arraybackslash}X@{}}
        \toprule
        \textbf{Condition} & \textbf{Class-conditioned} & \textbf{Unconditional} & \textbf{Unconditional} & \textbf{Camera control} \\
        Dataset & UCF-101 & UCF-101 & FFS / Sky / Taichi & RealEstate10K \\
        \midrule
        Guidance & CFG or CFG+RG & RG & None & HG \\
        Guidance scale & CFG 4, RG 0.6 & 0.8 & - & 4\\
        Solver & Euler & Heun & Heun & Euler \\
        Steps & \multicolumn{4}{c}{50} \\
        Noise shift $s$ & \multicolumn{4}{c}{4} \\
        FVD samples & \multicolumn{4}{c}{2,048} \\
        \bottomrule
    \end{tabularx}
\end{table}

\begin{table}[htp]
    \centering
    \caption{\textbf{Complete JVT-XL denoiser efficiency results on one NVIDIA 4090 GPU.} Embedding FLOPs count the convolution operations in the patch embedder. The embedding share is their percentage of denoiser FLOPs. Peak denotes allocated GPU memory. $C_b$ is the bottleneck width.}
    \label{tab:efficiency-denoiser-full}
    \scriptsize
    \setlength{\tabcolsep}{3pt}
    \renewcommand{\arraystretch}{1.08}
    \resizebox{\linewidth}{!}{
        \begin{tabular}{@{}llccccccc@{}}
            \toprule
            \multirow{2}{*}{\textbf{Embedding}} & \multirow{2}{*}{$C_b$} &
            \multirow{2}{*}{\textbf{In-context}} & \multirow{2}{*}{\textbf{Tokens}} &
            \multirow{2}{*}{\textbf{Params (M)}} & \multicolumn{3}{c}{\textbf{FLOPs}} &
            \multirow{2}{*}{\textbf{Peak (GiB)}} \\
            \cmidrule(lr){6-8}
            & & & & & \textbf{Total (TFLOPs)} & \textbf{Embed. (GFLOPs)} &
            \textbf{Embed. share (\%)} & \\
            \midrule
            Tubelet & -- & yes & 1056 & 682.2 & 1.089 & 7.248 & 0.666 & 2.674 \\
            Tubelet & 32 & yes & 1056 & 679.2 & 1.083 & 1.107 & 0.102 & 2.650 \\
            Tubelet & 64 & yes & 1056 & 679.7 & 1.084 & 2.215 & 0.204 & 2.652 \\
            Tubelet & 128 & yes & 1056 & 680.8 & 1.086 & 4.429 & 0.408 & 2.659 \\
            Patch-first & 16 & yes & 1056 & 678.7 & 1.082 & 0.252 & 0.023 & 2.648 \\
            Patch-first & 32 & yes & 1056 & 678.8 & 1.082 & 0.503 & 0.047 & 2.648 \\
            Patch-first & 64 & yes & 1056 & 679.0 & 1.082 & 1.007 & 0.093 & 2.648 \\
            Frame-first & 8 & yes & 1056 & 681.0 & 1.086 & 4.882 & 0.449 & 2.658 \\
            Frame-first & 16 & yes & 1056 & 683.3 & 1.091 & 9.764 & 0.895 & 2.673 \\
            Frame-first & 32 & yes & 1056 & 688.1 & 1.101 & 19.529 & 1.774 & 2.721 \\
            \midrule
            Tubelet & -- & no & 1024 & 682.1 & 1.063 & 7.248 & 0.682 & 2.674 \\
            Tubelet & 32 & no & 1024 & 679.1 & 1.057 & 1.107 & 0.105 & 2.649 \\
            Tubelet & 64 & no & 1024 & 679.7 & 1.058 & 2.215 & 0.209 & 2.651 \\
            Patch-first & 32 & no & 1024 & 678.8 & 1.057 & 0.503 & 0.048 & 2.648 \\
            Frame-first & 32 & no & 1024 & 688.0 & 1.076 & 19.529 & 1.815 & 2.721 \\
            \bottomrule
        \end{tabular}
    }
\end{table}

\subsection{Efficiency Evaluation Details}
\label{sec:appendix_efficiency_results}

The efficiency experiments use JVT-XL with 28 Transformer blocks, width 1,152, 16 attention heads, and spatial patch size 16. Each case processes one 16-frame $256\times256$ video in pixel space. With a temporal downsampling factor of 4, the denoiser receives 1,024 visual tokens. The register-enabled configuration inserts 32 in-context register tokens after the first 9 blocks, resulting in 1,056 tokens for the remaining blocks.

Sampling uses Euler integration with 50 steps. Register-enabled cases include no guidance, CFG, RG, and CFG+RG, while register-disabled cases include no guidance and CFG. These four guidance settings use one, two, two, and three denoiser-forward equivalents per step, respectively.

All measurements use bfloat16 with TF32 enabled and \texttt{torch.compile} disabled on one NVIDIA 4090 GPU. Each case runs in a fresh Python process. Denoiser latency uses 20 warm-up iterations followed by 100 timed iterations, and sampling latency uses 5 warm-up samples followed by 20 timed samples. The complete matrix is repeated in three independent sweeps. We record model parameters, FLOPs, baseline and peak allocated GPU memory, incremental peak memory, latency statistics, and throughput. FLOPs for one denoiser forward pass are counted with \texttt{torch.utils.flop\_counter.FlopCounterMode}. Sampling FLOPs are computed from the matching denoiser FLOPs, 50 sampling steps, and the number of denoiser-forward equivalents per step.

Tab.~\ref{tab:efficiency-denoiser-full} covers the ten embedding configurations in Tab.~\ref{tab:patch-embedding-bottleneck}, with additional register-free measurements for five of them. Tab.~\ref{tab:efficiency-sampling-full} reports the corresponding 30 sampling measurements. 

\begin{table}[htbp]
    \centering
    \caption{\textbf{Complete 50-step sampling efficiency results on one NVIDIA 4090 GPU.} ``Fwd.'' is the number of denoiser-forward equivalents per step. Latency is milliseconds per generated $16\times256\times256$ video and is reported as mean $\pm$ standard deviation across three sweeps. FLOPs are nominal estimates based on 50 steps and the listed forward equivalents. Register-free branch costs are approximated by the listed configuration's denoiser cost.}
    \label{tab:efficiency-sampling-full}
    \scriptsize
    \setlength{\tabcolsep}{2.6pt}
    \renewcommand{\arraystretch}{1.03}
    \resizebox{\linewidth}{!}{
        \begin{tabular}{@{}lcccccc@{}}
            \toprule
            \textbf{Embedding} & \textbf{In-context} & \textbf{Guidance} &
            \textbf{Fwd.} & \textbf{Latency (ms)} &
            \textbf{Peak (GiB)} & \textbf{Est. FLOPs (TFLOPs)} \\
            \midrule
            Tubelet (plain) & yes & None & 1 & $802.3\pm1.3$ & 2.689 & 54.4 \\
            Tubelet (plain) & yes & CFG & 2 & $1280.8\pm1.9$ & 2.751 & 108.9 \\
            Tubelet (plain) & yes & RG & 2 & $1529.9\pm0.3$ & 2.707 & 108.9 \\
            Tubelet (plain) & yes & CFG+RG & 3 & $2029.3\pm2.5$ & 2.768 & 163.3 \\
            Tubelet (plain) & no & None & 1 & $739.3\pm0.8$ & 2.689 & 53.2 \\
            Tubelet (plain) & no & CFG & 2 & $1190.5\pm2.4$ & 2.751 & 106.3 \\
            \midrule
            Tubelet $C_b=32$ & yes & None & 1 & $788.2\pm0.9$ & 2.662 & 54.1 \\
            Tubelet $C_b=32$ & yes & CFG & 2 & $1229.7\pm1.9$ & 2.724 & 108.3 \\
            Tubelet $C_b=32$ & yes & RG & 2 & $1544.9\pm1.9$ & 2.683 & 108.3 \\
            Tubelet $C_b=32$ & yes & CFG+RG & 3 & $1992.3\pm1.6$ & 2.747 & 162.4 \\
            Tubelet $C_b=32$ & no & None & 1 & $743.9\pm2.1$ & 2.662 & 52.9 \\
            Tubelet $C_b=32$ & no & CFG & 2 & $1168.5\pm1.1$ & 2.723 & 105.7 \\
            \midrule
            Tubelet $C_b=64$ & yes & None & 1 & $790.3\pm1.9$ & 2.663 & 54.2 \\
            Tubelet $C_b=64$ & yes & CFG & 2 & $1238.7\pm0.7$ & 2.726 & 108.4 \\
            Tubelet $C_b=64$ & yes & RG & 2 & $1532.5\pm1.7$ & 2.685 & 108.4 \\
            Tubelet $C_b=64$ & yes & CFG+RG & 3 & $1978.1\pm3.3$ & 2.749 & 162.5 \\
            Tubelet $C_b=64$ & no & None & 1 & $729.9\pm1.6$ & 2.663 & 52.9 \\
            Tubelet $C_b=64$ & no & CFG & 2 & $1158.2\pm1.1$ & 2.726 & 105.8 \\
            \midrule
            Patch-first $C_b=32$ & yes & None & 1 & $767.8\pm1.5$ & 2.661 & 54.1 \\
            Patch-first $C_b=32$ & yes & CFG & 2 & $1199.6\pm1.6$ & 2.721 & 108.2 \\
            Patch-first $C_b=32$ & yes & RG & 2 & $1498.1\pm1.3$ & 2.682 & 108.2 \\
            Patch-first $C_b=32$ & yes & CFG+RG & 3 & $1937.0\pm1.2$ & 2.746 & 162.3 \\
            Patch-first $C_b=32$ & no & None & 1 & $717.8\pm0.4$ & 2.660 & 52.8 \\
            Patch-first $C_b=32$ & no & CFG & 2 & $1144.0\pm1.1$ & 2.721 & 105.7 \\
            \midrule
            Frame-first $C_b=32$ & yes & None & 1 & $796.8\pm1.5$ & 2.737 & 55.0 \\
            Frame-first $C_b=32$ & yes & CFG & 2 & $1245.8\pm2.1$ & 2.781 & 110.1 \\
            Frame-first $C_b=32$ & yes & RG & 2 & $1507.7\pm2.1$ & 2.755 & 110.1 \\
            Frame-first $C_b=32$ & yes & CFG+RG & 3 & $1964.4\pm3.3$ & 2.799 & 165.1 \\
            Frame-first $C_b=32$ & no & None & 1 & $721.6\pm1.2$ & 2.737 & 53.8 \\
            Frame-first $C_b=32$ & no & CFG & 2 & $1152.5\pm0.3$ & 2.780 & 107.6 \\
            \bottomrule
        \end{tabular}
    }
\end{table}

\subsection{Additional Ablations}\label{sec:appendix_ablations}

Figure~\ref{fig:fvd-loss-btnk-dim-c2v} gives the bottleneck width sweep curve. In this experiment, we use batch size of 40, different from the full-scale experiments in Tab.~\ref{tab:patch-embedding-bottleneck} with global batch size 256. For tubelet and patch-first embedding, FVD varies non-monotonically with $C_b$: an overly narrow bottleneck may discard task-relevant spatiotemporal information, whereas an excessively wide bottleneck offers limited additional benefit. As for frame-first embedding, FVD improves progressively as $C_b$ increases, but the efficiency metrics declines sharply.

\begin{table}[htbp]
    \centering
    \caption{\textbf{Register-token and sampling ablation for unconditional video generation.} We report FVD@2048, lower is better. NFE denotes the number of function evaluations, and No reg. denotes the register-free baseline.}
    \label{tab:register-uncond}
    \small
    \setlength{\tabcolsep}{4pt}
    \renewcommand{\arraystretch}{1.15}
    \resizebox{\linewidth}{!}{
        \begin{tabular}{@{}cccccccccccccccc@{}}
            \toprule
            \multirow{2.5}{*}{\textbf{RG}}
            & \multirow{2.5}{*}{\textbf{NFE}}
            & \multirow{2.5}{*}{\textbf{Solver}}
            & \multirow{2.5}{*}{\textbf{Steps}} 
            & \multicolumn{3}{c}{\textbf{UCF-101}}
            & \multicolumn{3}{c}{\textbf{FaceForensics}}
            & \multicolumn{3}{c}{\textbf{SkyTimelapse}}
            & \multicolumn{3}{c}{\textbf{Taichi-HD}} \\
            \cmidrule(lr){5-7} \cmidrule(lr){8-10} \cmidrule(lr){11-13} \cmidrule(lr){14-16}
            & & &
            & No reg. & $e_r=0$ & $e_r=e_t$
            & No reg. & $e_r=0$ & $e_r=e_t$
            & No reg. & $e_r=0$ & $e_r=e_t$
            & No reg. & $e_r=0$ & $e_r=e_t$ \\
            \midrule
            \xmark
            & 100 & Heun & 50
            & \underline{413.5} & 378.6 & 474.2
            & \textbf{32.9} & \underline{37.7} & \underline{33.1}
            & \underline{46.6} & 54.9 & 52.6
            & 119.1 & 144.6 & 127.1 \\
            \xmark
            & 200 & Euler & 200
            & 421.2 & 392.4 & 484.5
            & 35.2 & 40.2 & 34.9 
            & 49.0 & 58.3 & 55.3
            & \underline{114.5} & 143.6 & \underline{124.3} \\
            \xmark
            & 200 & Heun & 100
            & \textbf{387.7} & \underline{354.9} & \textbf{451.5}
            & \underline{34.0} & 39.0 & 34.0
            & \textbf{45.5} & \underline{54.0} & \underline{51.2}
            & \textbf{114.4} & \underline{137.5} & \textbf{122.0} \\
            \cmark
            & 200 & Heun & 50
            & -- & \textbf{215.9} & \underline{462.4}
            & -- & \textbf{34.9} & \textbf{31.8}
            & -- & \textbf{47.8} & \textbf{51.1}
            & -- & \textbf{132.3} & \underline{124.3} \\
            \bottomrule
        \end{tabular}
    }
\end{table}

Tab.~\ref{tab:register-uncond} and Tab.~\ref{tab:adaln-ablation-uncond} repeat the register and AdaLN ablations of Tab.~\ref{tab:register-c2v} and Tab.~\ref{tab:adaln-ablation} (Sec.~\ref{sec:ablation_study}) for unconditional video generation. As shown in Tab.~\ref{tab:register-uncond}, the gains from registers are dataset-dependent: register guidance substantially improves UCF-101 but brings marginal or no improvement on FaceForensics, SkyTimelapse and Taichi-HD. The AdaLN ablation in Tab.~\ref{tab:adaln-ablation-uncond} leads to the same conclusion as in the class-conditioned case: simply removing AdaLN without scaling parameters commonly degrades FVD, supporting standard AdaLN as the default conditioning mechanism.

\begin{table}[htbp]
    \centering
    \caption{\textbf{Effect of timestep conditioning and adaptive layer normalization on unconditional video generation.} We report FVD@2048, where lower is better. For the AdaLN condition and $e_r$, -- denotes the absence of the corresponding conditioning pathway.}
    \label{tab:adaln-ablation-uncond}
    \small
    \setlength{\tabcolsep}{4pt}
    \begin{tabular}{@{}lccccc@{}}
            \toprule
            \textbf{Variant} & \textbf{AdaLN condition} & $\boldsymbol{e_r}$ & \textbf{FaceForensics} & \textbf{SkyTimelapse} & \textbf{Taichi-HD} \\
            \midrule
            baseline               & $e_t$ & --    & 32.9 & 46.6 & 119.1 \\
            w/ $e_r=e_t$           & $e_t$ & $e_t$ & 33.1 & 53.6 & 127.1 \\
            w/o AdaLN              & --    & --    & 40.6 & 60.8 & 118.2 \\
            \bottomrule
    \end{tabular}
\end{table}

Prior work on text-to-image generation has shown that, under matched parameter counts, AdaLN-free variants can outperform their standard AdaLN counterparts~\citep{wang2026minit2i}. However, such a parameter-matched comparison is not compute-matched: AdaLN accounts for a considerable fraction of the model parameters, yet it is merely a linear layer with linear computational complexity. Matching the parameter count of an AdaLN-free variant to the baseline therefore substantially enlarges the quadratic-complexity attention blocks, leading to a marked increase in overall computation. For this reason, the AdaLN-free variants in Tab.~\ref{tab:adaln-ablation} and Tab.~\ref{tab:adaln-ablation-uncond} are instead compared against standard JVT with the same number of Transformer layers.

\begin{table}[htp]
    \centering
    \caption{\textbf{Efficiency comparison of standard JVT and its AdaLN-free variant.} We experiment on JVT for class-conditioned task. -- denotes removal of AdaLN or register tokens.}
    \label{tab:adaln-efficiency}
    \small
    \resizebox{\linewidth}{!}{
        \begin{tabular}{@{}lccccccc@{}}
            \toprule
            \textbf{Variant} & \textbf{AdaLN condition} & $\boldsymbol{e_r}$ & \textbf{Depth} & \textbf{Params (M)} & \textbf{FLOPs (T)} & \textbf{Latency (ms)} & \textbf{Peak (GiB)} \\
            \midrule
            baseline     & $e_t+e_y$ & --    & 28 & 678.8 & 1.057 & 14.55 & 2.678 \\
            w/ $e_r=e_y$ & $e_t+e_y$ & $e_y$ & 28 & 678.8 & 1.082 & 15.66 & 2.648 \\
            w/o AdaLN    & -- & $e_y$ & 28 & 451.4 & 1.081 & 13.59 & 1.794 \\
            w/o AdaLN    & -- & $e_y$ & 42 & 674.6 & 1.624 & 20.12 & 2.650 \\
            \bottomrule
        \end{tabular}
    }
\end{table}

\subsection{Additional FVD Results}
\label{sec:appendix_fvd}

Training strategies and algorithm designs greatly affect model performance. Some methods adopt the image-video joint training strategy proposed in VDM~\citep{ho2022vdm}, which markedly improves video generation quality and can also be incorporated into our method. Tab.~\ref{tab:fvd-c2v-img} compares our method without the joint training strategy against these methods. We further compare our method with recent hybrid approaches of autoregressive modeling and diffusion, and show the results in Tab.~\ref{tab:fvd-c2v-ardiffusion}. 

\begin{table}[htp]
    \centering
    \caption{\textbf{FVD@2048 on different datasets.} We report class-conditioned generation on UCF-101 and unconditional generation on other datasets. ``+IMG" means video-image joint training.}
    \label{tab:fvd-c2v-img}
    \small
    \begin{tabular}{@{}lcccc@{}}
        \toprule
        \textbf{Method} &
        \textbf{UCF101} &
        \textbf{FaceForensics} &
        \textbf{SkyTimelapse} &
        \textbf{Taichi-HD} \\
        \midrule
        Latte~\citep{ma2025latte} & 477.97 & 34.00 & 59.82 & 159.6 \\
        Latte+IMG~\citep{ma2025latte} & 202.23 & 27.08 & 42.67 & 97.1 \\
        FrameDiT+IMG~\citep{le2026framedit} & 170.1 & 16.6 & 39.5 & 95.5 \\
        \textbf{JVT (Ours)} & 136.5 & 32.86 & 46.58 & 119.1 \\
        \bottomrule
    \end{tabular}
\end{table}

\begin{table}[htp]
    \centering
    \caption{\textbf{FVD@2048 on UCF-101.} We compare class-conditioned and unconditional generation with hybrid AR+diffusion methods. All methods use $16\times256\times256$ resolution.}
    \label{tab:fvd-c2v-ardiffusion}
    \small
    \begin{tabular}{@{}lcc@{}}
        \toprule
        \textbf{Method} & \textbf{FVD w/ class} & \textbf{FVD w/o class}\\
        \midrule
        AR-Diffusion~\citep{sun2025ardiffusion} & 186.6 & - \\
        Ca2-VDM~\citep{gao2025ca2vdm} & 184.5 &  - \\
        ACDiT-XL~\citep{hu2026acdit} & 111 & - \\
        MAGI~\citep{zhou2025magi} & - & 297.8 \\
        FAR-XL~\citep{gu2025far} & 108 & 279 \\
        JVT (Ours) & 136.5 & 215.9\\
        \bottomrule
    \end{tabular}
\end{table}

\begin{wraptable}{r}{0.52\textwidth}
    \vspace{-1.5em}
    \centering
    \caption{\textbf{FVD@10000 on UCF-101.} All methods generate 16-frame videos use class condition.}
    \label{tab:fvd-c2v-10000}
    \small
    \begin{tabular*}{\linewidth}{@{}lcc@{}}
        \toprule
        \textbf{Method} & \textbf{Res.} & \textbf{FVD@10000} \\
        \midrule
        MAGVIT~\citep{yu2023magvit} & $128^2$ & 76 \\
        MAGVIT-v2~\citep{yu2024magvitv2} & $128^2$ & 58 \\
        MaGNeTS~\citep{goyal2025magnets} & $128^2$ & 96.4 \\
        PAR~\citep{wang2025par} & $128^2$ & 94.1 \\
        ELT~\citep{goyal2026elt} & $128^2$ & 60.8 \\
        JVT (Ours) & $256^2$& 91.2 \\
        \bottomrule
    \end{tabular*}
\end{wraptable}

Different evaluation protocols can also directly influence the reported FVD scores. Following StyleGAN-V~\citep{skorokhodov2022styleganv} and Latte~\citep{ma2025latte}, we compute FVD using 2048 generated video samples (FVD@2048), whereas several other methods use 10000 samples (FVD@10000). Since the original FVD paper~\citep{unterthiner2019fvd} shows that the number of samples used for evaluation can greatly affect the FVD score, we additionally report our FVD@10000 result in Tab.~\ref{tab:fvd-c2v-10000}.

\clearpage

\begin{figure}[!htbp]
    \centering
    \resizebox{!}{0.92\textheight}{\input{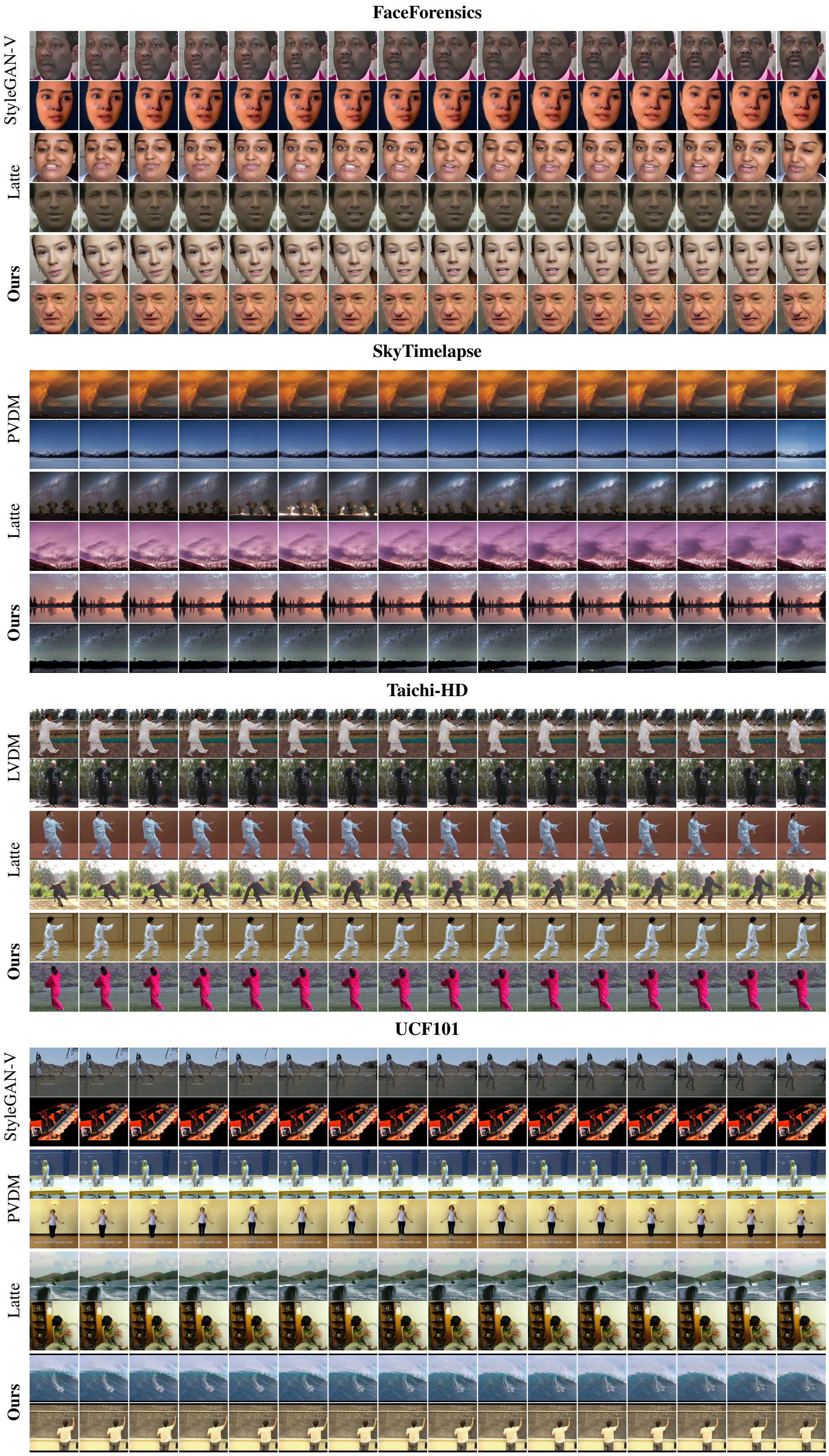}}
    \caption{\textbf{Video generation results across four datasets.} Two 16-frame samples are shown for each method on FaceForensics, SkyTimelapse, Taichi-HD, and UCF-101.}
    \label{fig:appendix-uncond-comparison}
\end{figure}

\end{document}